\documentclass[11pt]{article}

\usepackage[final]{acl}

\usepackage{times}
\usepackage{latexsym}

\usepackage[T1]{fontenc}

\usepackage[utf8]{inputenc}

\usepackage{microtype}

\usepackage{inconsolata}

\usepackage{graphicx}
\usepackage{booktabs}
\usepackage{amssymb}
\usepackage{multirow}
\usepackage{amsmath}
\usepackage[all]{hypcap}

\title{When Models Hear What They Expect: Diagnosing Prosodic Heuristics in Multimodal Sarcasm Detection}

\author{
  \textbf{Yongjian Chen}$^{1,2}$,
  \textbf{Pengfei Wei}$^{1}$,
  \textbf{Yiqun Sun}$^{1}$\thanks{Corresponding author},
  \textbf{Zhu Li}$^{2}$,
  \textbf{Lawrence B. Hsieh}$^{1}$ \\
  $^{1}$Magellan Technology Research Institute (MTRI) \\
  \texttt{\{pengfei.wei, duke.sun, lawrence.hsieh\}@mtri.co.jp} \\
    $^{2}$Center for Language and Cognition, University of Groningen, the Netherlands \\
    \texttt{\{yongjian.chenl, zhu.li\}@rug.nl} \\
}

\begin{document}
\maketitle
\begin{abstract}
Multimodal Large Language Models (MLLMs) process speech and text
jointly, yet whether they exploit prosodic cues for pragmatic
inference or rely on surface acoustic patterns has received
little systematic investigation. We address this through sarcasm
detection, evaluating Qwen2.5-Omni and Qwen3-Omni on Mandarin
Chinese and English under five modality conditions that decompose
the contributions of lexical content, vocal semantics, and
prosodic structure. Adding audio systematically inflates false
positives without improving true positive detection. Acoustic
error diagnosis reveals that model errors cluster on a shared
stereotype of expressive prosody, namely elevated pitch and
irregular pausing, which diverges from the actual cues marking
sarcasm in both languages. Targeted manipulation of only these
two dimensions causally confirms the heuristic, inducing false
positive rates of up to 60\%. Applying the same manipulation
template to Gemini~3 Flash Preview without modification
replicates the effect, 
suggesting that the stereotype extends beyond the Qwen Omni family rather than arising from a single model architecture.
\end{abstract}

\section{Introduction}
\label{sec:intro}

Sarcasm in speech is not just a matter of what you say but how you say
it. Speakers slow down, flatten or raise their pitch, and shift their
voice quality in ways that signal the gap between literal and intended
meaning. This prosodic signal has been documented across many languages,
with slower speech rate emerging as one of the most consistent
cross-linguistic markers~\citep{cheang2009,lan2025cantonese}. In
Mandarin, creaky voice and lower fundamental frequency (F0) are additional defining
features~\citep{gu2020voice,chen2020mandarin}. 
However, some prosodic cues do not generalise straightforwardly across or even within languages.
Fundamental frequency direction is a striking case: across studies of
English alone, some find sarcasm produced with a lower mean F0
\citep{rockwell2000,cheang2008,chen2018}, while others find the
opposite~\citep{anolli2002,gonzalez2016,jansen2020}. The same instability extends across languages: in Cantonese, \citet{cheang2009} report elevated mean F0 for sarcasm, whereas \citet{lan2025cantonese}, using more naturalistic elicitation with native Hong Kong speakers, find the opposite. As \citet{tatar2026prosody} note, this
inconsistency persists even after controlling for methodology. Despite
this variability, human listeners can recognise sarcasm from prosodic
cues alone, including from the very opening of an utterance before any
sarcasm-bearing word has been
heard~\citep{tatar2026prosody,mauchand2021}. Utterance duration emerges
as the most robust perceptual cue, while F0 direction remains
speaker- and context-dependent.

Recent MLLMs such as SALMONN~\citep{tang2023salmonn} and 
Qwen-Audio~\citep{chu2023qwenaudio} have shown competitive 
results on tasks from speech recognition~\citep{radford-etal-2023-whisper, tang2024contextualized} to spoken language 
understanding~\citep{peng2024survey, tang2023squad}. Yet on tasks requiring 
deeper paralinguistic reasoning, a consistent pattern 
emerges: the audio channel loses out when text is present. 
In spoken question answering, \citet{chi2025role} find that 
models do reasonably well on prosody-only input, but once 
text is available they largely ignore the acoustic channel. 
In speech emotion recognition, \citet{corrêa2025emotion} 
show that when vocal expression and word meaning point in 
opposite directions, models follow the words. 
\citet{wang2026emotionthinker} address this directly, 
building a training set augmented with prosodic annotations 
and a reinforcement learning scheme to push models toward 
acoustic reasoning; without this intervention, models default 
to lexical shortcuts. 

Sarcasm is precisely the case where lexical content and acoustic delivery can be in tension, so a model that ignores prosody is missing
the defining signal of the phenomenon \cite{gao2025jest, sun2026position}. Prior computational work has
mostly treated sarcasm as a text and vision problem~\citep{gao2025jest},
and evaluation of MLLMs on spoken sarcasm is only just beginning.
\citet{gao2025jest} survey the field and note the near-absence of
cross-lingual work and the underexplored role of speech data.
\citet{liu2025evaluating} evaluate MLLMs on English and Chinese sarcasm corpora under multiple modality conditions and show that performance varies with the available inputs. However, their study primarily benchmarks modality-conditioned recognition performance, but it does not identify which acoustic properties drive model errors or test whether such properties are causally responsible for those errors. Our study addresses this diagnostic gap by moving from modality-conditioned performance to acoustic error analysis and causal intervention.


We address this gap by evaluating two MLLMs on Mandarin Chinese (MSCD) and English (MUStARD++) sarcasm corpora. We design five modality conditions that systematically decompose the contributions of lexical content, vocal semantics, and prosodic structure (\S\ref{sec:setup}). We then extract 66 acoustic features from both corpora to establish language-specific prosodic ground truths for sarcasm (\S\ref{sec:acoustic-ground-truth}), diagnose model errors against these ground truths (\S\ref{sec:model-cues}), and causally verify the identified heuristic through targeted prosodic manipulation (\S\ref{sec:causal}). 

\begin{figure*}[t]
    \centering
    \includegraphics[width=\textwidth]{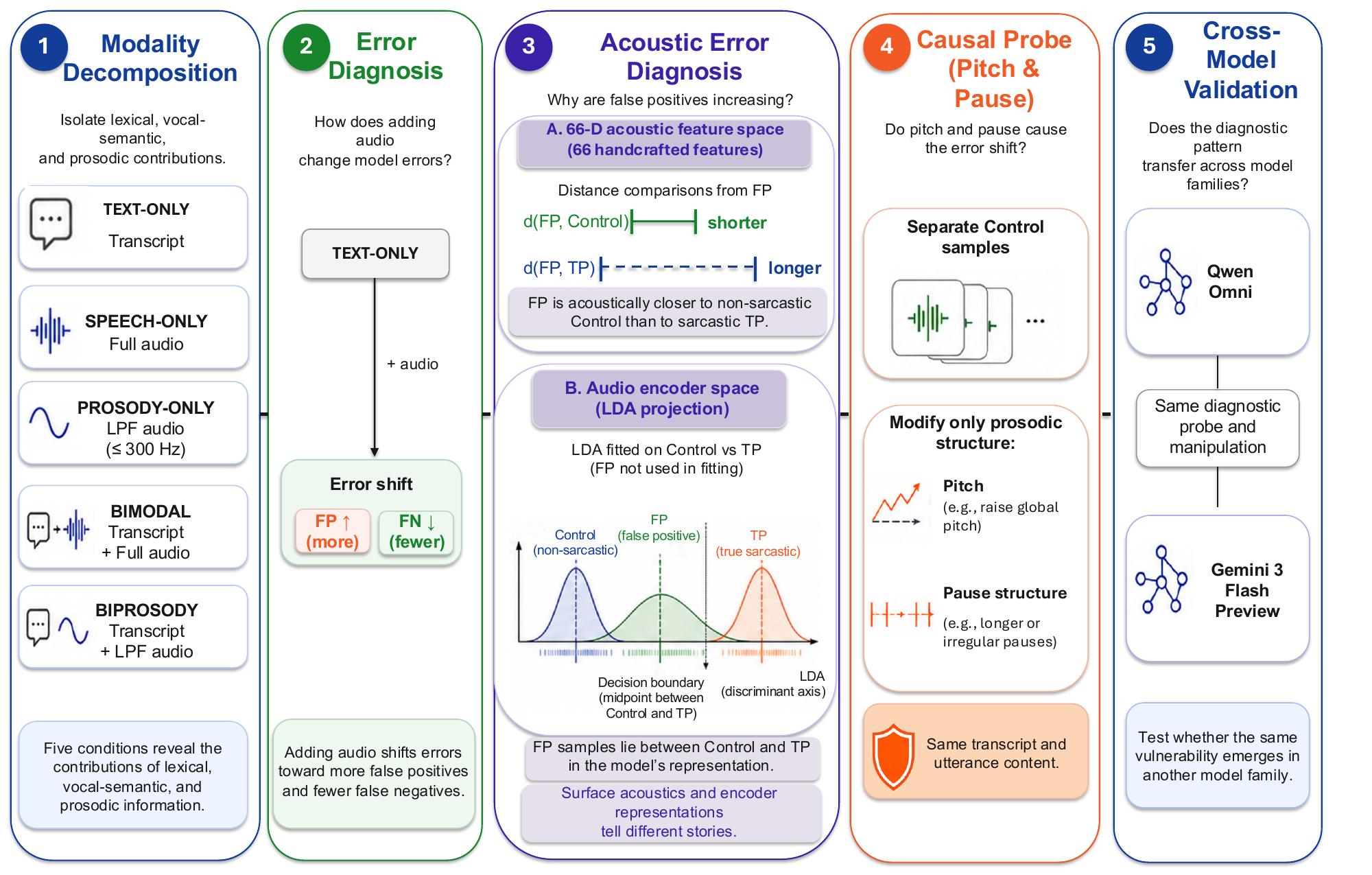}
    \caption{Overview of the diagnostic framework for multimodal sarcasm
    detection, from modality decomposition to causal validation.}
    \label{fig:overview}
\end{figure*}

Our main contributions are as follows: 
(1)~a diagnostic evaluation framework using five modality conditions to separate lexical, vocal-semantic, and prosodic contributions, applied to two MLLMs across English and Mandarin Chinese;
(2)~the first acoustic characterisation of
false-positive errors in multimodal sarcasm detection,
identifying a cross-linguistic expressive prosody stereotype
diverging from corpus-derived sarcasm signatures; 
(3)~causal
validation via targeted PSOLA manipulation, inducing false
positive rates of up to 60\% by modifying only pitch and pause
structure.
Figure~\ref{fig:overview} provides an overview of our diagnostic
framework. We first decompose lexical, vocal-semantic, and prosodic
contributions across five modality conditions, then trace the resulting false-positive shift to acoustic error profiles. We next test the identified prosodic dimensions causally through targeted manipulation on separate Control samples, and examine whether the same diagnostic pattern transfers beyond the Qwen Omni family.

\section{Experimental Setup}
\label{sec:setup}

\subsection{Datasets}
\label{sec:datasets}

We evaluate on two multimodal sarcasm corpora from typologically distinct languages.
\textbf{MSCD} \cite{liu-etal-2022-mscd} is a Chinese corpus of televised stand-up comedy, providing 2,705 samples balanced at 49.7\% sarcasm. \textbf{MUStARD++} \cite{yue-etal-2019-mmsd} is an English corpus drawn from sit-coms, providing 1,202 audio-transcript pairs with near-perfect class balance (50.1\% sarcasm). Both corpora feature authentic conversational speech with studio audience reactions, which motivates the audio enhancement step described in Section~\ref{sec:audio}. Table~\ref{tab:datasets} summarises the splits used throughout this work. 
Since all models are evaluated zero-shot, the standard train/validation/test splits do not have a learning-theoretic role in our experiments. We retain the original splits to facilitate comparison with future work that uses these partitions for training. We further partition the training split into five folds and evaluate each fold independently, using the folds as repeated
evaluation runs over the larger partition rather than as cross-validation for model fitting. This provides a more stable estimate of performance variance and allows us to verify that the observed modality and error patterns are not driven by a particular subset of the data. We report mean $\pm$ standard deviation across the five folds together with the fixed validation and test evaluations.

\begin{table}[t]
\centering
\small
\begin{tabular}{lccc}
\toprule
\textbf{Dataset} & \textbf{Train} & \textbf{Validation} & \textbf{Test} \\
\midrule
MUStARD++ ($n$=1,202) & 841 & 180 & 181 \\
MSCD ($n$=2,705) & 1,893 & 406 & 406 \\
\bottomrule
\end{tabular}
\caption{Dataset statistics. Train sets are partitioned into 5 folds
(MUStARD++: 168--169 per fold; MSCD: 378--379 per fold). Both corpora are
near-perfectly balanced ($\approx$\,50\% sarcasm).}
\label{tab:datasets}
\vspace{-6mm}
\end{table}

\subsection{Models}
\label{sec:models}

We evaluate two models from the Qwen Omni family zero-shot:
\textit{Qwen2.5-Omni-7B} \cite{xu-etal-2025-qwen2omni}, referred
to as \textit{Omni~2.5}, and
\textit{Qwen3-Omni-30B-A3B-Thinking} \cite{xu-etal-2025-qwen3omni},
referred to as \textit{Omni~3.0}. We select this family for two
reasons: both models are openly available, and both accept
text-only input without requiring a paired audio signal, a
prerequisite for the \textsc{Text-only} condition in our design.
Both share a Thinker--Talker architecture that processes audio
natively without cascading through automatic speech recognition(ASR), but differ in audio
encoding capacity. Omni~2.5 uses a Whisper-based encoder
\cite{radford-etal-2023-whisper} projected into a 7B dense
language model backbone. Omni~3.0 replaces this with a
purpose-built Audio Transformer encoder trained from scratch
on 20 million hours of supervised audio, coupled with a 30B
Mixture-of-Experts backbone. Omni~3.0 is used in thinking mode,
producing explicit chain-of-thought traces analysed as secondary
evidence in \S\ref{sec:modality:fp} and \S\ref{sec:causal:results}.
For cross-architecture generalisation, we additionally evaluate
\textit{Gemini~3 Flash Preview} \cite{google2025gemini3pro},
a closed-source model from Google DeepMind, in a targeted
transfer analysis (\S\ref{sec:causal:transfer}).

\subsection{Modality Conditions}
\label{sec:conditions}




We define five modality conditions that systematically decompose
the contributions of lexical content, vocal semantics, and
prosodic structure. The three unimodal conditions isolate
individual channels. \textsc{Text-only} provides the textual semantic baseline, \textsc{Speech-only} supplies the full voice signal without text, and \textsc{Prosody-only} retains only prosodic structure by reducing lexical content from the audio. The two multimodal conditions both pair the transcript with audio but
differ in what that audio carries. \textsc{Bimodal} uses the
full voice signal while \textsc{Biprosody} uses filtered audio.
Because the transcript is held constant across both conditions, any performance gap reflects the contribution of vocal semantics rather than artefacts of the distribution shift introduced by low-pass filtering.

\subsection{Audio Preprocessing}
\label{sec:audio}

All audio conditions use speech-enhanced recordings; the prosody-filtered conditions (\textsc{Prosody-only} and \textsc{Biprosody}) additionally apply low-pass filtering to remove lexical content from the signal.

\paragraph{Speech enhancement.}
Both corpora contain studio audience laughter and background noise that introduce spurious cues and degrade the prosodic signal. We compared two enhancement models, FRCRN\_SE\_16K \citep{zhao-etal-2022-frcrn} and MossFormerGAN\_SE\_16K \citep{zhao-etal-2022-mossformers} against the original audio using DNSMOS P.835 \citep{reddy-etal-2022-dnsmos}, P808 \citep{naderi-etal-2020-p808}, and WVMOS \citep{andreev-etal-2022-wvmos}. MossFormerGAN outperformed FRCRN across all metrics for both languages and was selected for all subsequent processing. Full metric scores, pairwise significance tests, and raw vs.\ enhanced downstream F1 comparisons are reported in Appendix~\ref{app:enhancement}.

\paragraph{Low-pass filtering.}
The \textsc{Prosody-only} and \textsc{Biprosody} conditions require audio from which lexical content has been removed while prosodic structure is preserved. Following \citet{chi2025role}, we apply a low-pass filter at 300\,Hz, a cutoff empirically validated to retain F0 contours, intensity patterns, and rhythmic structure while removing the high-frequency spectral detail that carries phonemic and word-level identity. The filter is applied to the enhanced audio rather than the original recordings, so that the prosodic signal is not confounded by residual laughter or background noise.

To verify lexical suppression, we transcribe enhanced full-speech and
\textsc{Prosody-only} audio with ASR. CER is 9.01\% (EN) and 13.61\% (ZH) for full speech, versus 91.11\% and 153.74\% for
\textsc{Prosody-only}. The Chinese CER above 100\% reflects spurious insertions. These results show that word-level information is largely unrecoverable from the low-pass-filtered signal.

\subsection{Acoustic Feature Extraction}
\label{sec:features}

To support the prosodic analyses in later sections, we extract acoustic features from the enhanced audio using Praat~\cite{boersma-weenink-2021-praat} via the Parselmouth Python interface~\cite{jadoul-etal-2018-parselmouth}. The 66 features fall into four categories: fundamental frequency (20), intensity and energy (17), rhythm and timing (15), and voice quality (14). The full feature list is provided in Appendix~\ref{app:features}.

\subsection{Implementation Details}
\label{sec:implementation}

All experiments are conducted using the ms-swift framework
\cite{zhao-etal-2024-swift} with a vLLM backend \cite{kwon-etal-2023-vllm},
running on two NVIDIA H100 GPUs.
Models are prompted with a binary classification instruction to output
\texttt{true} or \texttt{false} to indicate the presence of sarcasm.
Three prompt templates are used: text-only for
\textsc{Text-only}, audio-only shared by \textsc{Speech-only}
and \textsc{Prosody-only}, and audio-with-transcript shared by
\textsc{Bimodal} and \textsc{Biprosody}. All templates share
the same binary classification instruction, ensuring that
performance differences within each condition pair are
attributable solely to input content. Full templates are
provided in Appendix~\ref{app:prompts}.

 

\section{What Audio Changes: Performance and Error Shifts}
\label{sec:modality}

Introducing audio does not uniformly improve sarcasm detection. Instead, it shifts the model's error profile, trading missed sarcasm for false alarms. We report F1 performance across all five modality conditions (Figure~\ref{fig:f1_dotplot}) and directional error analysis across four audio--text contrasts
(Figure~\ref{fig:scatter_fp_fn}), with per-split breakdowns for both in Appendix~\ref{app:splits}.

\subsection{Overall Performance by Modality Condition}
\label{sec:modality:f1}

\begin{figure}[t]
  \centering
  \includegraphics[width=\columnwidth]{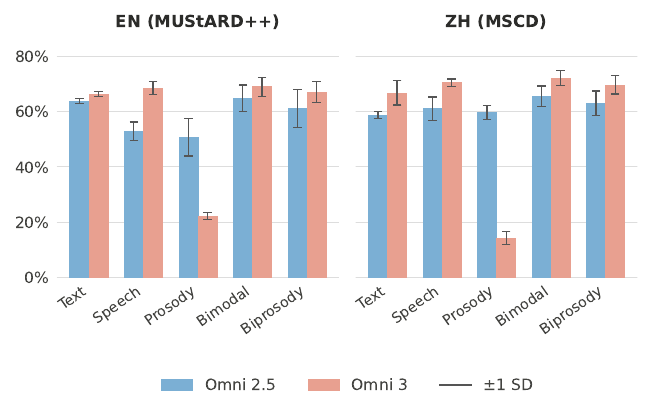}
    \caption{Mean F1 (\%) by modality condition across train
    (5-fold average), validation, and test splits. Error bars
    show $\pm$1~SD across splits; tighter bars indicate greater
    cross-split stability. \textsc{Prosody-only} is included to
    establish that isolated prosody is non-functional for this
    task; it is excluded from the error analysis in
    \S\ref{sec:modality:fp}.}
      
  \label{fig:f1_dotplot}
  \vspace{-4mm}
\end{figure}

Two observations stand out from Figure~\ref{fig:f1_dotplot}.

\paragraph{\textsc{Prosody-only} collapses.}
For \textbf{Omni~3.0}, \textsc{Prosody-only} falls dramatically
below all other conditions in both languages: 22.2\,\% mean F1
in EN and 14.3\,\% in ZH, well below the 50\,\% chance
baseline. The failure mode is catastrophic conservatism: the
model predicts sarcasm on almost no \textsc{Prosody-only}
utterances, yielding near-zero recall. For \textbf{Omni~2.5},
\textsc{Prosody-only} approaches chance in EN (50.7\,\%) but
appears inflated in ZH (59.6\,\%), a precision-recall artefact
driven by high recall at near-random precision rather than
genuine discriminative ability. In both cases the evaluated
models show no reliable discriminative ability under this
condition, and we therefore exclude \textsc{Prosody-only}
from the error analysis that follows.

\paragraph{Three trends are visible across the functional conditions.}
\textsc{Bimodal} is consistently the top-ranked or joint-top
condition across all four model--language panels, with mean gains
of $+$1.0 to $+$6.8\,pp over \textsc{Text-only} across splits.
\textsc{Biprosody} trails \textsc{Bimodal} by only 1.0--3.7\,pp
despite receiving filtered rather than full audio, suggesting that
vocal semantics contribute marginally to overall F1 and that
prosodic contour carries most of the audio effect.
The benefit of audio is larger in Chinese than English. \textbf{Omni~3.0} gains $+$5.3\,pp in ZH but only $+$2.7\,pp
in EN for \textsc{Bimodal}, which we attribute to the
informationally richer role of prosody in a tonal language, where
vocal acoustics carry lexical content unavailable from the
transcript alone. \textsc{Speech-only} diverges sharply between
models. \textbf{Omni~3.0} matches or exceeds \textsc{Text-only}
in both languages ($+$3.7\,pp in ZH, $+$2.2\,pp in EN), whereas
\textbf{Omni~2.5} falls $-$10.9\,pp below \textsc{Text-only}
in EN and is near-flat in ZH ($+$2.4\,pp), reflecting the
limited prosodic extraction capacity of its Whisper-based encoder
relative to the purpose-built Audio Transformer in
\textbf{Omni~3.0} (see \S\ref{sec:models}).

\subsection{Audio Shifts Errors: The False Positive Signature}
\label{sec:modality:fp}

\begin{figure*}[t]
  \centering
  \includegraphics[width=\textwidth]{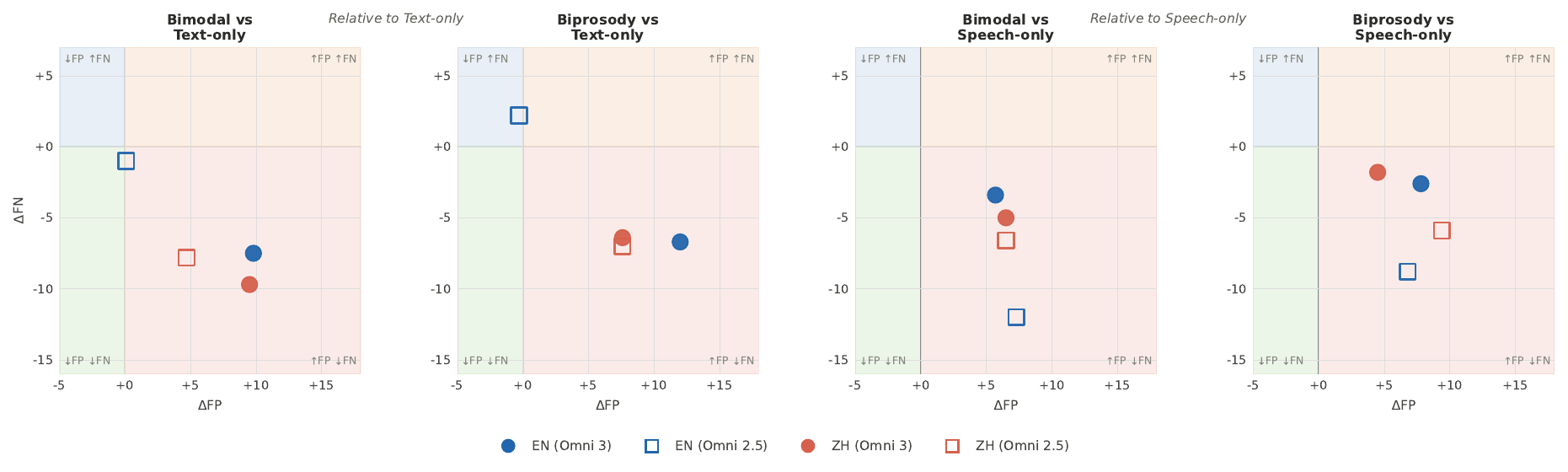}
    \caption{$\Delta$FP vs.\ $\Delta$FN for four modality
    contrasts, grouped by baseline: \textsc{Text-only} (left
    two panels) and \textsc{Speech-only} (right two panels).
    Values are expressed as percentage of total samples; each
    point represents the mean across train, validation, and
    test splits. Quadrant shading indicates the direction of
    error change: lower-right (coral) = $\uparrow$FP
    $\downarrow$FN; upper-left (blue) = $\downarrow$FP
    $\uparrow$FN; lower-left (green) = $\downarrow$FP
    $\downarrow$FN; upper-right (orange) = $\uparrow$FP
    $\uparrow$FN. Color encodes language (blue = EN, red = ZH);
    fill encodes model (filled = \textbf{Omni~3.0}, hollow =
    \textbf{Omni~2.5}).}
  \label{fig:scatter_fp_fn}
\end{figure*}

While F1 captures overall performance trends, it obscures how
errors change across conditions. Figure~\ref{fig:scatter_fp_fn}
reveals that the F1 gains observed in \S\ref{sec:modality:f1}
co-occur with a systematic directional bias, whereby adding
audio trades false negatives for false positives.
We examine this from two baselines, using \textsc{Text-only}
to isolate what audio adds to a transcript, and
\textsc{Speech-only} to isolate what a transcript adds to audio.

\paragraph{\textsc{Bimodal} and \textsc{Biprosody} inflate FPs
relative to \textsc{Text-Only}.}
The left two panels show that both audio conditions push
Omni~3.0 consistently into the lower-right quadrant across
both languages. \textsc{Bimodal} adds $+$9.8\,\% FP in EN
and $+$8.6\,\% in ZH (mean across splits), with corresponding
FN reductions of $-$7.5\,\% and $-$9.7\,\%;
\textsc{Biprosody} produces nearly identical inflation
(EN: $+$12.0\,\%, ZH: $+$7.6\,\%). For Omni~2.5, the same
direction holds in ZH (\textsc{Bimodal}: $+$4.7\,\%;
\textsc{Biprosody}: $+$7.6\,\%), while EN shows negligible
shifts ($+$0.1\,\% and $-$0.3\,\% respectively), clustering
near the origin.

\paragraph{The transcript activates the mismatch heuristic.}
The right two panels reveal the complementary pattern: adding
a transcript on top of either full speech or prosody-only audio
pushes points into the lower-right quadrant across both models
and languages, adding 1--12\,\% of samples as additional false
positives. The driving mechanism is the co-presence of a literal
transcript and an audio signal: the transcript anchors the verbal
content as non-sarcastic, while the audio introduces a perceived
tonal contrast that the model interprets as ironic mismatch.
Inspection of chain-of-thought traces reveals a characteristic
reasoning pattern among these false positives. In a representative
case, the \textsc{Bimodal} trace describes the audio as having a
``high-pitched, drawn-out tone'' that creates a ``clear
contradiction between words and tone''; the \textsc{Biprosody}
trace, operating on prosody-filtered audio without lexical content,
arrives at the same conclusion, calling the vocal delivery a
``playful exaggeration'' that ``completely contradicts'' the
literal transcript (full traces in Appendix~\ref{app:cot-examples}).
The convergence across both conditions indicates that the model
grounds its mismatch judgment in prosodic contour rather than
vocal semantics. Neither the transcript alone nor the audio alone
triggers the false alarm; it is their co-presence that does.

\section{Corpus-Derived Acoustic Profile of Sarcasm}
\label{sec:acoustic-ground-truth}

Before exploring what models hear, we investigate what sarcasm sounds like in the two corpora. We compare sarcastic and non-sarcastic utterances using Mann--Whitney~$U$ with Cohen's~$d$ as the primary effect size, cross-validated across the train, validation, and test partitions. Features are ranked by mean $|d|$ across splits. The goal is to characterise language-specific acoustic profiles and establish 
corpus-derived
cue directions against which model heuristics (§\ref{sec:model-cues}) can be assessed. The ranked prosodic profiles are in Appendix~\ref{app:prosody-ranking}.

\paragraph{Chinese (MSCD).}
The dominant acoustic signature of Mandarin sarcasm is pitch-structural and temporal. The leading features by mean $|d|$ are total pause duration ($d = 0.80$), utterance duration ($d = 0.76$), pitch contour complexity (f0\_num\_peaks: $d = 0.74$; f0\_num\_valleys: $d = 0.74$), and intensity structure (intensity\_num\_peaks: $d = 0.73$), with rhythmic regularity inverted ($d = -0.61$, sarcastic speech \emph{less} regular). All top features replicate across all three splits; pause\_duration\_total reaches $d = 0.90$ in the test partition, the only large effect in the analysis. The duration effect replicates the most cross-linguistically stable finding in the sarcasm prosody literature~\citep{cheang2009,lan2025cantonese,tatar2026prosody}. More distinctive is the dominance of F0 \emph{contour complexity} over F0 \emph{level}: sarcastic Mandarin is marked by substantially more pitch peaks and valleys ($d \approx 0.74$) while F0 mean does not appear among the top discriminating features.

\paragraph{English (MUStARD++).}
The English profile is structurally distinct. The strongest signals are energy-based, with
\textsc{intensity\_max} ($d = 0.47$), \textsc{rms\_range}
($d = 0.43$), \textsc{rms\_max} ($d = 0.43$), and
\textsc{rms\_std} ($d = 0.41$) as the leading features,
followed by a lower F0 across mean, median, and minimum
(f0\_median: $d = -0.40$; f0\_mean: $d = -0.36$;
f0\_min: $d = -0.30$) and longer duration ($d = 0.32$). The lower F0 finding replicates the dominant characterisation of English sarcasm~\citep{cheang2008,chen2018}, and the duration effect again confirms the cross-linguistic pattern~\citep{tatar2026prosody}.

The two profiles diverge most sharply on F0 direction: Chinese
sarcasm is dominated by pitch contour complexity (not F0 level),
while English sarcasm is consistently lower across all three F0
summary statistics (f0\_mean, f0\_median, f0\_min:
$d = -0.30$ to $-0.40\downarrow$), an inversion that holds
across all splits. This cross-linguistic divergence is consistent
with the broader literature showing that F0 mean direction is
not a stable cross-linguistic sarcasm cue~\citep{tatar2026prosody},
while duration effects generalise (\S\ref{sec:intro}).


\section{Error Diagnosis: What Do Models Actually Respond To?}
\label{sec:model-cues}

Section~\ref{sec:modality:fp} shows that prosodic input
systematically inflates false positives across models and languages,
but leaves open which acoustic properties drive the effect.
We diagnose this through two analyses: two pairwise distance tests
\footnote{Prior to distance computation, all acoustic features are standardised to zero mean and unit variance via \texttt{StandardScaler} fitted on the pooled FP, Control, and TP feature matrices. Centroid distances are computed between group mean vectors in this standardised space.} 
and a feature-profile comparison
(\S\ref{sec:fp-profile}), operating over three groups held constant
across ZH and EN:
\textbf{FP} utterances are non-sarcastic samples classified correctly from text alone, but flagged as sarcastic under both \textsc{Biprosody} and \textsc{Bimodal} conditions.
\textbf{TP}~are sarcastic utterances correctly classified across all
conditions.
\textbf{Control}~are non-sarcastic utterances correctly classified under both the \textsc{Biprosody} and \textsc{Bimodal} conditions.

\subsection{False Positives Are Not Acoustically Near True Sarcasm, but Remain Ambiguous to the Model}
\label{sec:fp-distance}

Table~\ref{tab:fp-distance} shows that false-positive utterances sit much 
closer to non-sarcastic controls than to true sarcasm in the hand-crafted 
feature space: the FP$\leftrightarrow$TP centroid distance exceeds 
FP$\leftrightarrow$Control by 2.35$\times$ (EN) and 3.23$\times$ (ZH) 
under Euclidean distance, and by 6.68$\times$ and 20.11$\times$ under 
cosine distance. Measured against corpus acoustics, false positives do 
not resemble the sarcasm they are mistaken for, and every FP sample 
individually assigns to the Control centroid in both languages.

\begin{figure*}[t]
    \centering
    \includegraphics[width=\textwidth, height=0.4\textheight, keepaspectratio]{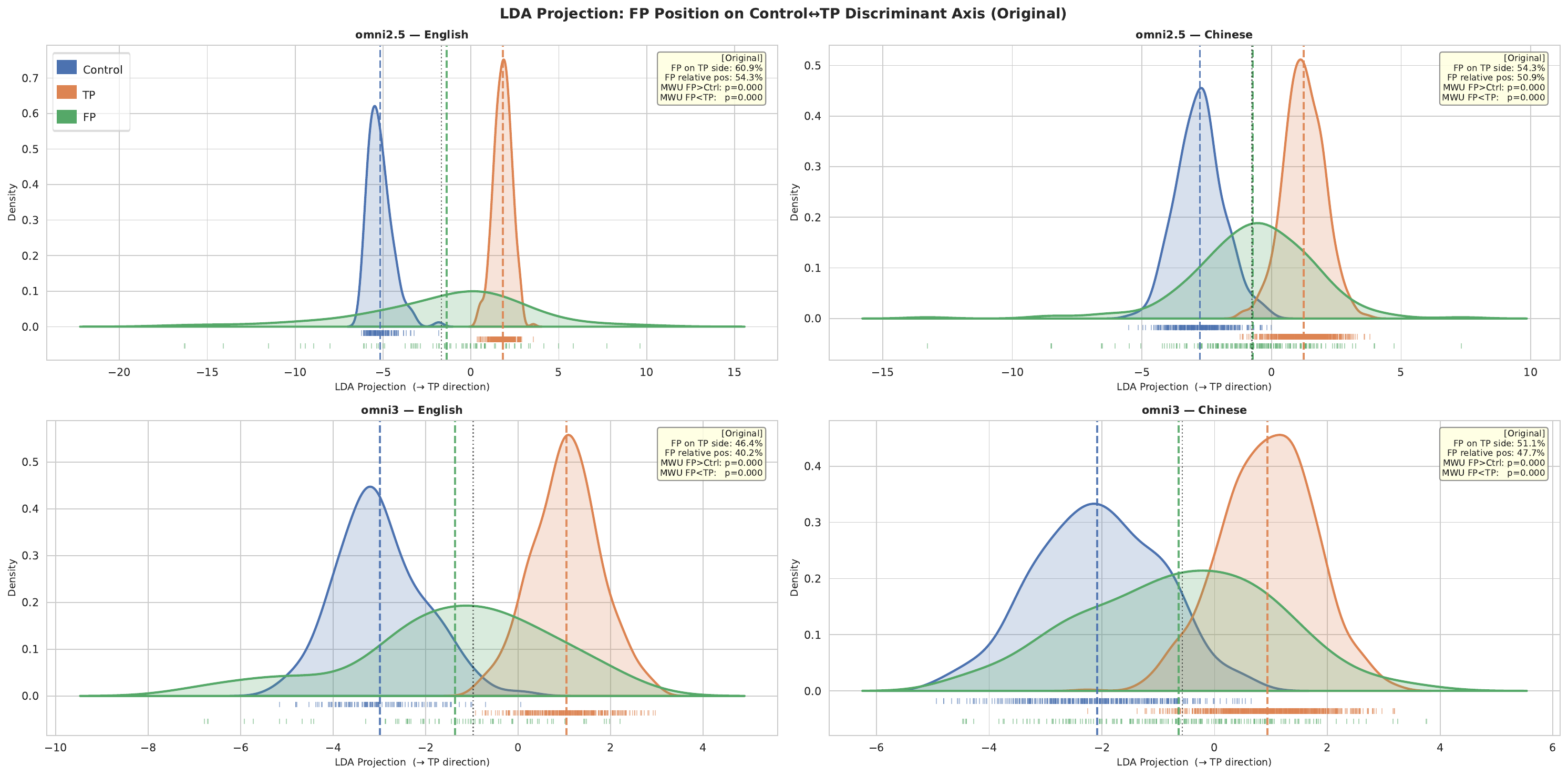}
    \caption{LDA projection of audio encoder embeddings onto the
    Control--TP discriminant axis. FP samples consistently fall between Control and TP across all four model--language conditions, although their handcrafted acoustic features are closer to Control. Dashed lines mark group means, and the dotted line marks the midpoint between
    Control and TP.}
    \label{fig:lda}
\end{figure*}

The model's audio encoder, however, does not draw the same
boundary. Projecting audio encoder embeddings onto the LDA
axis fitted to separate Control from TP representations
(with FP samples excluded from fitting) places FP
significantly between the two groups across all four
model--language conditions (Mann-Whitney $U$, $p < .001$;
Figure~\ref{fig:lda}).\footnote{LDA was fitted with shrinkage
regularisation to address high-dimensional embeddings.
Robustness to class imbalance was resolved via
minority-class oversampling.} FP representations hover near
the midpoint of the Control$\leftrightarrow$TP discriminant
axis, with substantial overlap across all three
distributions. Omni~2.5 places FP slightly closer to the
sarcastic pole, while Omni~3.0 pulls them marginally toward
Control. In neither case does the encoder commit these
utterances clearly to either side.

The same utterances that pattern unambiguously with Control in acoustic feature space are represented as intermediate by the audio encoder, occupying a region between sarcastic and non-sarcastic speech in the encoder representation space. This indicates that the ambiguity underlying the false-positive errors is already present at the audio encoding stage, although these results do not determine whether it is further amplified during multimodal fusion. Section~\ref{sec:fp-profile} asks which acoustic properties produce this internal ambiguity.

\begin{table}[t]
\centering
\small

\begin{tabular}{llccc}
\toprule
\textbf{Lang.} & \textbf{Metric} & $d(\text{FP}{\leftrightarrow}\text{Ctrl})$ & $d(\text{FP}{\leftrightarrow}\text{TP})$ & \textbf{Ratio} \\
\midrule
\multirow{2}{*}{EN} & Euclidean & 0.952 & 2.238 & 2.352 \\
                    & Cosine    & 0.275 & 1.834 & 6.675 \\
\midrule
\multirow{2}{*}{ZH} & Euclidean & 0.861 & 2.784 & 3.232 \\
                    & Cosine    & 0.097 & 1.951 & 20.112 \\
\bottomrule
\end{tabular}
\caption{Centroid distances and ratios
$d(\text{FP}{\leftrightarrow}\text{TP})\,/\,d(\text{FP}{\leftrightarrow}\text{Control})$; values $>1$ indicate FP is closer to Control.
Sample counts: EN $n_{\text{fp}}=69$, $n_{\text{ctrl}}=137$, $n_{\text{tp}}=352$;
ZH $n_{\text{fp}}=188$, $n_{\text{ctrl}}=391$, $n_{\text{tp}}=805$.}
\label{tab:fp-distance}
\vspace{-6mm}
\end{table}

\subsection{Feature Profiles of False Positives vs. Controls}
\label{sec:fp-profile}

Having shown that false positives are acoustically dissimilar from true sarcasm, we characterize their prosodic profile to identify what cues the model responds to instead (full FP--Control rankings in Appendix~\ref{app:fp-profile}).

\paragraph{Chinese (ZH).} The top ground-truth discriminators 
(\textsc{pause\_duration\_total}, \textsc{duration}, 
\textsc{f0\_num\_peaks}) capture sustained temporal and melodic 
complexity, none of which drive FP--Control separation. Instead, the 
model responds to elevated mean F0 ($d = 0.375$) and irregular pausing 
(\textsc{pause\_duration\_std} $d = 0.362$; \textsc{pause\_duration\_max} 
$d = 0.355$), where ground-truth pause effects reflect total duration 
rather than distributional irregularity.

\paragraph{English (EN).} EN yields only four significant FP--Control 
separators, a sparser profile than ZH, yet the mismatch is sharper. The 
ground-truth signature is intensity-prominent and F0-suppressed, whereas 
all four FP--Control separators show elevated F0 
(\textsc{F0\_max} $d = 0.268$; \textsc{F0\_range} $d = 0.255$) and 
increased pausing (\textsc{pause\_rate} $d = 0.223$; 
\textsc{num\_pauses} $d = 0.215$), positive where ground-truth F0 
effects are negative.

Across both languages, model errors cluster on elevated pitch and irregular pausing. In Chinese, this captures surface properties of the same acoustic domains that mark sarcasm (pitch, pausing) but targets the wrong aspects, e.g. pitch level rather than contour complexity, pause irregularity rather than total duration. In English, the mismatch is sharper: the FP profile is directionally opposite to the ground-truth signature, with elevated F0 where sarcasm exhibits suppressed F0. In both cases, the model responds to a cross-linguistic stereotype of expressive speech rather than authentic sarcastic cues.  
These findings motivate the causal manipulation in \S\ref{sec:causal}, which targets pitch level and pause structure because these are the dimensions on which false positives consistently diverge from controls.

\section{Causal Verification}
\label{sec:causal}

Section~\ref{sec:model-cues} identifies the acoustic profiles that
drive false positives; we now test whether these profiles are
causally linked to model errors by directly manipulating Control
samples to exhibit the FP signature.

\subsection{Targeted Prosodic Manipulation}
\label{sec:causal:design}

Manipulations are derived from the FP--Control acoustic profiles identified in \S\ref{sec:fp-profile} and applied to a separate set of Control samples that the models classify correctly under both \textsc{Bimodal} and \textsc{Biprosody} conditions. The samples used to derive the heuristic and those used for causal manipulation are non-overlapping. Thus, the manipulation constitutes an out-of-sample test. Rather than modifying the false-positive utterances from which the acoustic profile was identified, we impose the
identified prosodic pattern on previously correctly classified
non-sarcastic utterances and test whether the same error is induced. 

Manipulations are adapted to the phonological properties of each
language (Appendix~\ref{app:manipulation}). For \textbf{ZH}, the
four features with the largest FP--Control effect sizes are
targeted, spanning both F0 and rhythm: a uniform PSOLA shift of
$+8.8\%$ raises mean F0, combined with targeted pause elongation
to increase both maximum pause duration and distributional
irregularity. For \textbf{EN}, all four significant FP--Control
separators are targeted: the same PSOLA factor elevates F0, with
existing pauses stretched by $\times1.49$ to produce longer,
more salient pauses. Pause insertion was avoided in both languages to preserve naturalness. Manipulation magnitudes are also bounded by the 75th percentile of the FP feature distribution, the upper quartile of naturally occurring FP prosody, ensuring that manipulated samples remain within an empirically observed range rather than producing artificially extreme signals. \footnote{To verify intelligibility is preserved, we applied
ASR-based CER and DNSMOS to the manipulated stimuli. CER
differences are negligible for Chinese (0.084 vs.\ 0.079;
bootstrap 95\,\% CI crosses zero) and minimal for English
(increase of 0.017). The weak sample-level
$\Delta$CER--$\Delta$OVRL correlation ($|r| < 0.15$) provides no evidence that perceptual quality changes systematically impair intelligibility. Full results are reported in
Appendix~\ref{app:naturalness}.} The multi-cue logic is grounded in
human perception research: \citet{gonzalez2016} find that
simultaneous modification of pitch and duration in
sarcasm-consistent directions improves listener accuracy;
\citet{peters2016} similarly show that increased duration and
modified pitch lead English listeners to rate utterances as more
sarcastic. These findings provide a principled basis for expecting
analogous manipulations to influence model behaviour if models have
internalised similar perceptual associations.
\subsection{Results}
\label{sec:causal:results}

\begin{figure}[t]
\centering
\includegraphics[width=\columnwidth]{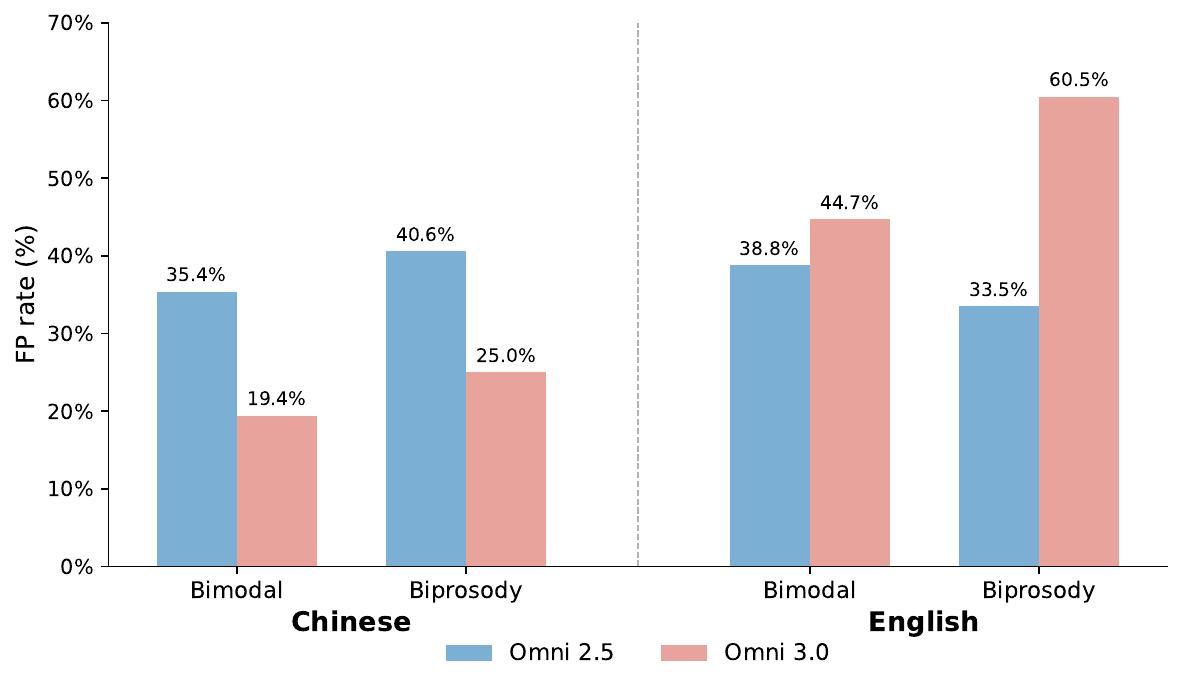}
\caption{FP rates after prosodic manipulation, computed as the
proportion of Control samples misclassified as sarcastic
following manipulation. ZH $n=268$; EN $n=95$.}
\label{fig:causal-results}
\vspace{-6mm}
\end{figure}

The manipulations produce substantial FP rate increases across
both languages and models (Figure~\ref{fig:causal-results}),
confirming that the acoustic profiles identified in
\S\ref{sec:fp-profile} are causally linked to model errors
rather than coincidental correlates.
ZH FP rates rise to 19.35--40.65\% depending on model and
condition; EN rates reach 33.55--60.53\%, with \textbf{Omni~3.0}
under \textsc{Biprosody} showing the highest attack-induced rate
across all conditions. These rates were achieved by modifying only two dimensions, i.e. pitch and pausing.

For \textbf{Omni~3.0}, a consistent biprosody $>$ bimodal ordering
emerges in both languages. We interpret this as a modality context
effect: in \textsc{Bimodal} conditions, the acoustic content of
full speech provides a grounding context against which manipulated
prosodic features are normalised, partially suppressing their
impact. In \textsc{Biprosody}, this grounding is absent. The model
receives only a prosody-only audio signal alongside the transcript,
and the contrast between the literal textual content and the
manipulated prosodic signal is rendered more salient, amplifying
the mismatch heuristic identified in \S\ref{sec:modality:fp}.
Chain-of-thought analysis on a representative ZH sample
illustrates the mechanism (full trace in
Appendix~\ref{app:cot-examples}): the same $+8.8\%$ F0 manipulation
is described as ``light, cheerful, amused'' under \textsc{Bimodal}, but as ``strained, high-pitched, forced laughter'' under
\textsc{Biprosody}, where the absence of vocal semantics causes
the model to interpret identical prosody as mocking rather than
expressive. This suggests that prosodic features do not carry
fixed perceptual weight for these models, but are interpreted
relative to the broader acoustic context in which they occur. Reverse manipulation, shifting FP samples toward control-like
prosodic targets, partially recovers correct classification
across both languages and models (29.5--56.3\,\% flip rates;
Appendix~\ref{app:reverse}), providing complementary
bidirectional evidence for the causal role of elevated F0
and irregular pausing.

\subsection{Model Transfer}
\label{sec:causal:transfer}

To test whether the identified heuristic generalises beyond the
Qwen Omni family, we evaluated Gemini~3 Flash Preview
\cite{google2025gemini3pro} on the same corpora and modality
conditions. The pattern mirrors that of the Omni models. Adding
audio inflates false positives relative to \textsc{Text-only}
across both languages and all three audio conditions, with
\textsc{Bimodal} and \textsc{Biprosody} producing the largest
$\Delta$FP in EN ($+$13.2\,pp and $+$13.6\,pp respectively),
replicating the core error signature without any
model-family-specific tuning (Table~\ref{tab:gemini}).

To further probe the causal role of pitch and pausing, we applied
the same manipulation template derived from the Omni FP profiles
(Appendix~\ref{app:manipulation}) to Gemini without modification.
The manipulation transfers across all three audio conditions and
both languages, with flip rates ranging from 4.7\,\% to
17.1\,\% depending on condition and language
(Table~\ref{tab:gemini}). That a template derived from a
different model family induces systematic misclassification in
Gemini confirms that the vulnerability to elevated pitch and
irregular pausing is not an artefact of the Qwen architecture.
Flip rates are lower than those observed for the Omni models,
consistent with differences in audio encoding capacity between
the two families.\footnote{For comparison, a Gemini-specific template derived from Gemini~3's own FP--Control geometry is also evaluated in Appendix~\ref{app:gemini}.}

\begin{table}[t]
\centering
\small
\setlength{\tabcolsep}{3.5pt}
\begin{tabular}{lrrrrrr}
\toprule
 & \multicolumn{3}{c}{\textbf{EN (MUStARD++)}} & \multicolumn{3}{c}{\textbf{ZH (MSCD)}} \\
\cmidrule(lr){2-4}\cmidrule(lr){5-7}
\textbf{Condition} & F1 & $\Delta$FP & Flip & F1 & $\Delta$FP & Flip \\
\midrule
\textsc{Text-only}   & 69.6 & ---   & ---  & 76.4 & ---  & ---  \\
\textsc{Speech-only} & 71.6 & +6.3  & 17.1 & 76.3 & +4.2 & 8.9  \\
\textsc{Bimodal}     & 74.7 & +13.2 & 14.3 & 77.0 & +4.6 & 5.0  \\
\textsc{Biprosody}   & 73.0 & +13.6 & 17.1 & 76.8 & +4.5 & 4.7  \\
\bottomrule
\end{tabular}
\caption{Gemini~3 Flash Preview: mean F1 (\%), $\Delta$FP
(percentage-point change relative to \textsc{Text-only}),
and FP flip rate (\%) under the Omni-derived manipulation
template across all three audio conditions.}
\label{tab:gemini}
\vspace{-4mm}
\end{table}

\section{Conclusion}
\label{sec:conclusion}

We evaluated two MLLMs on English and Mandarin Chinese sarcasm
corpora under five modality conditions and found that prosodic
audio systematically inflates false positives without improving
true positive detection. Acoustic error diagnosis reveals that
false positive utterances are not acoustically close to genuine
sarcasm. Instead, model errors track a language-independent
stereotype of expressive prosody, namely elevated pitch and
irregular pausing, that diverges from the actual sarcasm
signatures in both languages. Targeted prosodic manipulation
causally confirms this heuristic, inducing false positive rates
of up to 60\% by modifying only two acoustic dimensions, pitch
level and pause structure, neither of which aligns with the
corpus-derived prosodic signatures of sarcasm in either language.
The pattern holds across both models, both languages, and
Gemini~3 Flash Preview, a closed-source model from a
different developer, demonstrating that the shallow prosodic
stereotype is not a property of any single model family but
a broader failure mode of current MLLMs. These findings
suggest that acquiring genuine paralinguistic grounding will
require explicit alignment between prosodic representations
and language-specific pragmatic cues, rather than reliance
on surface acoustic correlations. These findings point to the value of targeted diagnostic evaluation that assesses whether models have internalised linguistically grounded prosodic knowledge, rather than relying on aggregate performance metrics alone.

\clearpage

\section*{Limitations}
While our results are consistent across model scales, languages,
and evaluation splits, three aspects of the current design leave
room for further investigation.

\paragraph{Corpus scope.}
Both MSCD and MUStARD++ consist of performative television speech. The acoustic profiles in \S\ref{sec:acoustic-ground-truth} therefore reflect the speech contexts represented in these benchmarks and may not fully generalise to spontaneous everyday interaction. We are not aware of a comparable natural-interaction corpus with paired audio and sarcasm labels. This limitation primarily concerns the external generality of the corpus-derived acoustic profiles, whereas the causal manipulation in \S\ref{sec:causal} is conducted within the same recordings with lexical content and recording context held fixed.

\paragraph{Training implications.}
Our findings identify elevated pitch and irregular pausing as the operative dimensions of a shallow prosodic stereotype, providing concrete targets for future training interventions. Contrastive augmentation pairing identical transcripts with prosodically distinct audio would force models to learn that surface acoustic patterns do not reliably signal pragmatic intent. Reinforcement learning objectives that reward sensitivity to language-specific prosodic cues over surface acoustic correlations are a complementary direction; \citet{wang2026emotionthinker} demonstrate that exactly this kind of prosody-aware \textsc{rl} intervention improves acoustic reasoning in speech emotion recognition without sacrificing text-based performance, suggesting the approach is viable for paralinguistic tasks more broadly.
\paragraph{Mechanism localisation.}
Our LDA analysis (\S\ref{sec:fp-distance}) shows that the audio encoder already places false positive utterances in an ambiguous region between sarcastic and non-sarcastic representations, but this characterises the encoder in isolation. Probing the intermediate stages of this interaction, where prosodic representations meet textual grounding, would clarify whether the bias is inherited from the encoder or constructed during fusion, and point to more targeted interventions.


\bibliography{custom}

\begin{thebibliography}{40}
\providecommand{\natexlab}[1]{#1}

\bibitem[{Andreev et~al.(2023)Andreev, Alanov, Ivanov, and Vetrov}]{andreev-etal-2022-wvmos}
Pavel Andreev, Aibek Alanov, Oleg Ivanov, and Dmitry~P. Vetrov. 2023.
\newblock {HIFI++}: A unified framework for bandwidth extension and speech enhancement.
\newblock In \emph{Proceedings of ICASSP 2023}, pages 1--5.

\bibitem[{Anolli et~al.(2002)Anolli, Ciceri, and Infantino}]{anolli2002}
Luigi Anolli, Rita Ciceri, and Maria~Giaele Infantino. 2002.
\newblock \href {https://doi.org/10.1080/00207590244000106} {From ``blame by praise'' to ``praise by blame'': Analysis of vocal patterns in ironic communication}.
\newblock \emph{International Journal of Psychology}, 37(5):266--276.

\bibitem[{Boersma and Weenink(2024)}]{boersma-weenink-2021-praat}
Paul Boersma and David Weenink. 2024.
\newblock Praat: Doing phonetics by computer.
\newblock Computer program.
\newblock Version 6.4, retrieved from \url{http://www.praat.org/}.

\bibitem[{Cheang and Pell(2008)}]{cheang2008}
Henry~S. Cheang and Marc~D. Pell. 2008.
\newblock \href {https://doi.org/10.1016/j.specom.2007.11.003} {The sound of sarcasm}.
\newblock \emph{Speech Communication}, 50(5):366--381.

\bibitem[{Cheang and Pell(2009)}]{cheang2009}
Henry~S. Cheang and Marc~D. Pell. 2009.
\newblock Acoustic markers of sarcasm in {Cantonese} and {English}.
\newblock \emph{The Journal of the Acoustical Society of America}, 126(3):1394--1405.

\bibitem[{Chen and Boves(2018)}]{chen2018}
Aoju Chen and Lou Boves. 2018.
\newblock \href {https://doi.org/10.1017/S0025100318000038} {What's in a word: Sounding sarcastic in {British English}}.
\newblock \emph{Journal of the International Phonetic Association}, 48(1):57--76.

\bibitem[{Chi et~al.(2025)Chi, de~Seyssel, and Schluter}]{chi2025role}
Jie Chi, Maureen de~Seyssel, and Natalie Schluter. 2025.
\newblock \href {https://doi.org/10.18653/v1/2025.findings-naacl.471} {The role of prosody in spoken question answering}.
\newblock In \emph{Findings of the Association for Computational Linguistics: NAACL 2025}, pages 8483--8494, Albuquerque, New Mexico. Association for Computational Linguistics.

\bibitem[{Chu et~al.(2023)Chu, Xu, Zhou, Yang, Zhang, Yan, Zhou, and Zhou}]{chu2023qwenaudio}
Yunfei Chu, Jin Xu, Xiaohuan Zhou, Qian Yang, Shiliang Zhang, Zhijie Yan, Chang Zhou, and Jingren Zhou. 2023.
\newblock \href {https://arxiv.org/abs/2311.07919} {{Qwen-Audio}: Advancing universal audio understanding via unified large-scale audio-language models}.
\newblock \emph{Preprint}, arXiv:2311.07919.

\bibitem[{Corr{\^e}a et~al.(2025)Corr{\^e}a, Lima, Moreno, Ueda, and Costa}]{corrêa2025emotion}
Pedro Corr{\^e}a, Jo{\~a}o Lima, Victor Moreno, Lucas Ueda, and Paula Dornhofer~Paro Costa. 2025.
\newblock \href {https://arxiv.org/abs/2510.25054} {Evaluating emotion recognition in spoken language models on emotionally incongruent speech}.
\newblock \emph{Preprint}, arXiv:2510.25054.

\bibitem[{Gao et~al.(2025{\natexlab{a}})Gao, Nayak, and Coler}]{gao2025jest}
Xiyuan Gao, Shekhar Nayak, and Matt Coler. 2025{\natexlab{a}}.
\newblock Spoken in jest, detected in earnest: A systematic review of sarcasm recognition---multimodal fusion, challenges, and future prospects.
\newblock \emph{IEEE Transactions on Affective Computing}, 16:2526--2544.

\bibitem[{Gao et~al.(2025{\natexlab{b}})Gao, Wang, Zhang, Huang, Li, Nayak, and Coler}]{liu-etal-2022-mscd}
Xiyuan Gao, Bruce~Xiao Wang, Meiling Zhang, Shuming Huang, Zhu Li, Shekhar Nayak, and Matt Coler. 2025{\natexlab{b}}.
\newblock \href {https://doi.org/10.21437/Interspeech.2025-1632} {A multimodal {Chinese} dataset for cross-lingual sarcasm detection}.
\newblock In \emph{Proceedings of Interspeech 2025}, pages 3968--3972.

\bibitem[{Geng et~al.(2021)Geng, Shi, and Guo}]{chen2020mandarin}
Puyang Geng, Shaopei Shi, and Hong Guo. 2021.
\newblock \href {https://doi.org/10.1145/3458380.3458414} {The coding strategy for the {Mandarin} speech conveying sarcasm in acoustic and articulatory domain}.
\newblock In \emph{Proceedings of the 2021 5th International Conference on Digital Signal Processing}, pages 195--200.

\bibitem[{Gonz{\'a}lez-Fuente et~al.(2016)Gonz{\'a}lez-Fuente, Prieto, and Noveck}]{gonzalez2016}
Santiago Gonz{\'a}lez-Fuente, Pilar Prieto, and Ira Noveck. 2016.
\newblock \href {https://doi.org/10.21437/SpeechProsody.2016-185} {A fine-grained analysis of the acoustic cues involved in verbal irony recognition in {French}}.
\newblock In \emph{Proceedings of Speech Prosody 2016}, pages 902--906.

\bibitem[{{Google DeepMind}(2025)}]{google2025gemini3pro}
{Google DeepMind}. 2025.
\newblock \href {https://storage.googleapis.com/deepmind-media/Model-Cards/Gemini-3-Pro-Model-Card.pdf} {Gemini 3 {P}ro model card}.

\bibitem[{Jadoul et~al.(2018)Jadoul, Thompson, and de~Boer}]{jadoul-etal-2018-parselmouth}
Yannick Jadoul, Bill~D. Thompson, and Bart~G. de~Boer. 2018.
\newblock Introducing {Parselmouth}: A {Python} interface to {Praat}.
\newblock \emph{Journal of Phonetics}, 71:1--15.

\bibitem[{Jansen and Chen(2020)}]{jansen2020}
Nelleke Jansen and Aoju Chen. 2020.
\newblock \href {https://doi.org/10.21437/SpeechProsody.2020-84} {Prosodic encoding of sarcasm at the sentence level in {Dutch}}.
\newblock In \emph{Proceedings of Speech Prosody 2020}, pages 409--413.

\bibitem[{Kwon et~al.(2023)Kwon, Li, Zhuang, Sheng, Zheng, Yu, Gonzalez, Zhang, and Stoica}]{kwon-etal-2023-vllm}
Woosuk Kwon, Zhuohan Li, Siyuan Zhuang, Ying Sheng, Lianmin Zheng, Cody~Hao Yu, Joseph Gonzalez, Hao Zhang, and Ion Stoica. 2023.
\newblock \href {https://doi.org/10.1145/3600006.3613165} {Efficient memory management for large language model serving with {PagedAttention}}.
\newblock In \emph{Proceedings of the 29th Symposium on Operating Systems Principles}, SOSP'23, pages 611--626.

\bibitem[{Lan and Mok(2025)}]{lan2025cantonese}
Chen Lan and Peggy Mok. 2025.
\newblock \href {https://doi.org/10.1177/00238309251333766} {Acoustic cues in the production and perception of {Cantonese} sarcasm}.
\newblock \emph{Language and Speech}.

\bibitem[{Li et~al.(2020)Li, Gu, Liu, and Tang}]{gu2020voice}
Shanpeng Li, Wentao Gu, Lei Liu, and Ping Tang. 2020.
\newblock \href {https://doi.org/10.1044/2020_JSLHR-19-00166} {The role of voice quality in {Mandarin} sarcastic speech: An acoustic and electroglottographic study}.
\newblock \emph{Journal of Speech, Language, and Hearing Research}, 63:2578--2588.

\bibitem[{Li et~al.(2025)Li, Gao, Zhang, Nayak, and Coler}]{liu2025evaluating}
Zhu Li, Xiyuan Gao, Yuqing Zhang, Shekhar Nayak, and Matt Coler. 2025.
\newblock \href {https://arxiv.org/abs/2509.15476} {Evaluating multimodal large language models on spoken sarcasm understanding}.
\newblock \emph{Preprint}, arXiv:2509.15476.

\bibitem[{Mauchand et~al.(2021)Mauchand, Caballero, Jiang, and Pell}]{mauchand2021}
Ma{\"e}l Mauchand, Jonathan~A. Caballero, Xiaoming Jiang, and Marc~D. Pell. 2021.
\newblock \href {https://doi.org/10.3758/s13415-020-00849-7} {Immediate online use of prosody reveals the ironic intentions of a speaker: Neurophysiological evidence}.
\newblock \emph{Cognitive, Affective, \& Behavioral Neuroscience}, 21(1):74--92.

\bibitem[{Naderi and Cutler(2020)}]{naderi-etal-2020-p808}
Babak Naderi and Ross Cutler. 2020.
\newblock An open source implementation of {ITU-T} recommendation {P.808} with validation.
\newblock In \emph{Proceedings of Interspeech 2020}.

\bibitem[{Peng et~al.(2024)Peng, Wang, Li, Guo, Wang, Fang, Xi, Li, Li, Zhang, Wang, and Yu}]{peng2024survey}
Jing Peng, Yucheng Wang, Bohan Li, Yiwei Guo, Hankun Wang, Yangui Fang, Yu~Xi, Haoyu Li, Xu~Li, Kexin Zhang, Shuai Wang, and Kai Yu. 2024.
\newblock A survey on speech large language models for understanding.
\newblock \emph{IEEE Journal of Selected Topics in Signal Processing}, 20:2--31.

\bibitem[{Peters et~al.(2016)Peters, Wilson, Boiteau, Gelormini-Lezama, and Almor}]{peters2016}
Sara Peters, Kathryn Wilson, Timothy~W. Boiteau, Carlos Gelormini-Lezama, and Amit Almor. 2016.
\newblock \href {https://doi.org/10.1017/S1366728915000048} {Do you hear it now? {A} native advantage for sarcasm processing}.
\newblock \emph{Bilingualism: Language and Cognition}, 19(2):400--414.

\bibitem[{Radford et~al.(2023)Radford, Kim, Xu, Brockman, McLeavey, and Sutskever}]{radford-etal-2023-whisper}
Alec Radford, Jong~Wook Kim, Tao Xu, Greg Brockman, Christine McLeavey, and Ilya Sutskever. 2023.
\newblock Robust speech recognition via large-scale weak supervision.
\newblock In \emph{Proceedings of the 40th International Conference on Machine Learning}, ICML'23.

\bibitem[{Ray et~al.(2022)Ray, Mishra, Nunna, and Bhattacharyya}]{yue-etal-2019-mmsd}
Anupama Ray, Shubham Mishra, Apoorva Nunna, and Pushpak Bhattacharyya. 2022.
\newblock A multimodal corpus for emotion recognition in sarcasm.
\newblock In \emph{Proceedings of the Thirteenth Language Resources and Evaluation Conference}, pages 6992--7003, Marseille, France. European Language Resources Association.

\bibitem[{Reddy et~al.(2022)Reddy, Gopal, and Cutler}]{reddy-etal-2022-dnsmos}
Chandan K.~A. Reddy, Vishak Gopal, and Ross Cutler. 2022.
\newblock {DNSMOS P.835}: A non-intrusive perceptual objective speech quality metric to evaluate noise suppressors.
\newblock In \emph{Proceedings of ICASSP 2022}, pages 886--890.

\bibitem[{Rockwell(2000)}]{rockwell2000}
Patricia Rockwell. 2000.
\newblock \href {https://doi.org/10.1023/A:1005120109296} {Lower, slower, louder: Vocal cues of sarcasm}.
\newblock \emph{Journal of Psycholinguistic Research}, 29(5):483--495.

\bibitem[{Sun et~al.(2025)Sun, Huang, Xu, Sun, Tang, and Tung}]{sun2025one}
Yandong Sun, Qiang Huang, Ziwei Xu, Yiqun Sun, Yixuan Tang, and Anthony~KH Tung. 2025.
\newblock One swallow does not make a summer: Understanding semantic structures in embedding spaces.
\newblock \emph{arXiv preprint arXiv:2512.00852}.

\bibitem[{Sun et~al.(2026)Sun, Huang, Tung, and Yu}]{sun2026position}
Yiqun Sun, Qiang Huang, Anthony Kum~Hoe Tung, and Jun Yu. 2026.
\newblock \href {https://openreview.net/forum?id=7NL8On4v6m} {Position: Text embeddings should capture implicit semantics, not just surface meaning}.
\newblock In \emph{Forty-third International Conference on Machine Learning Position Paper Track}.

\bibitem[{Tang et~al.(2023{\natexlab{a}})Tang, Yu, Sun, Chen, Tan, Li, Lu, Ma, and Zhang}]{tang2023salmonn}
Changli Tang, Wenyi Yu, Guangzhi Sun, Xianzhao Chen, Tian Tan, Wei Li, Lu~Lu, Zejun Ma, and Chao Zhang. 2023{\natexlab{a}}.
\newblock \href {https://arxiv.org/abs/2310.13289} {{SALMONN}: Towards generic hearing abilities for large language models}.
\newblock \emph{Preprint}, arXiv:2310.13289.

\bibitem[{Tang and Tung(2024)}]{tang2024contextualized}
Yixuan Tang and Anthony~KH Tung. 2024.
\newblock Contextualized speech recognition: Rethinking second-pass rescoring with generative large language models.
\newblock In \emph{IJCAI}, pages 6478--6485.

\bibitem[{Tang et~al.(2023{\natexlab{b}})Tang, Tung, and Elkind}]{tang2023squad}
Yixuan Tang, Anthony~KH Tung, and Edith Elkind. 2023{\natexlab{b}}.
\newblock Squad-src: A dataset for multi-accent spoken reading comprehension.
\newblock In \emph{IJCAI}, pages 5206--5214.

\bibitem[{Tat{\'a}r et~al.(2026)Tat{\'a}r, Brennan, Krivokapi{\'c}, and Keshet}]{tatar2026prosody}
Csilla Tat{\'a}r, Jonathan~R. Brennan, Jelena Krivokapi{\'c}, and Ezra Keshet. 2026.
\newblock \href {https://doi.org/10.1017/S002510032500009X} {Does prosody mark sarcasm early in an utterance? {A} production and perception study, including listeners who self-identified as being on the autism spectrum}.
\newblock \emph{Journal of the International Phonetic Association}, pages 1--40.

\bibitem[{Wang et~al.(2026)Wang, Liu, Zhang, Chen, Li, and Meng}]{wang2026emotionthinker}
Dingdong Wang, Shujie Liu, Tianhua Zhang, Youjun Chen, Jinyu Li, and Helen Meng. 2026.
\newblock Emotionthinker: Prosody-aware reinforcement learning for explainable speech emotion reasoning.
\newblock In \emph{International Conference on Learning Representations (ICLR)}.

\bibitem[{Xu et~al.(2025)Xu, Guo, He, Hu, He, Bai, Chen, Wang, Fan, Dang, Zhang, Wang, Chu, and Lin}]{xu-etal-2025-qwen2omni}
Jin Xu, Zhifang Guo, Jinzheng He, Hangrui Hu, Ting He, Shuai Bai, Keqin Chen, Jialin Wang, Yang Fan, Kai Dang, Bin Zhang, Xiong Wang, Yunfei Chu, and Junyang Lin. 2025.
\newblock \href {https://arxiv.org/abs/2503.20215} {{Qwen2.5-Omni} technical report}.
\newblock \emph{Preprint}, arXiv:2503.20215.

\bibitem[{Yang et~al.(2025)Yang, Li, Yang, Zhang, Hui, Zheng, Yu, Gao, Huang, Lv, Zheng, Liu, Zhou, Huang, Hu, Ge, Wei, Lin, Tang, Yang, Tu, Zhang, Yang, Yang, Zhou, Zhou, Lin, Dang, Bao, Yang, Yu, Deng, Li, Xue, Li, Zhang, Wang, Zhu, Men, Gao, Liu, Luo, Li, Tang, Yin, Ren, Wang, Zhang, Ren, Fan, Su, Zhang, Zhang, Wan, Liu, Wang, Cui, Zhang, Zhou, and Qiu}]{xu-etal-2025-qwen3omni}
An~Yang, Anfeng Li, Baosong Yang, Beichen Zhang, Binyuan Hui, Bo~Zheng, Bowen Yu, Chang Gao, Chengen Huang, Chenxu Lv, Chujie Zheng, Dayiheng Liu, Fan Zhou, Fei Huang, Feng Hu, Hao Ge, Haoran Wei, Huan Lin, Jialong Tang, and 41 others. 2025.
\newblock \href {https://arxiv.org/abs/2505.09388} {Qwen3 technical report}.
\newblock \emph{Preprint}, arXiv:2505.09388.

\bibitem[{Zhao and Ma(2023)}]{zhao-etal-2022-mossformers}
Shengkui Zhao and Bin Ma. 2023.
\newblock {MossFormer}: Pushing the performance limit of monaural speech separation using gated single-head transformer with convolution-augmented joint self-attentions.
\newblock In \emph{Proceedings of ICASSP 2023}, pages 1--5.

\bibitem[{Zhao et~al.(2022)Zhao, Ma, Watcharasupat, and Gan}]{zhao-etal-2022-frcrn}
Shengkui Zhao, Bin Ma, Karn~N. Watcharasupat, and W.~S. Gan. 2022.
\newblock {FRCRN}: Boosting feature representation using frequency recurrence for monaural speech enhancement.
\newblock In \emph{Proceedings of ICASSP 2022}, pages 9281--9285.

\bibitem[{Zhao et~al.(2025)Zhao, Huang, Hu, Wang, Mao, Zhang, Jiang, Wu, Ai, Wang, Zhou, and Chen}]{zhao-etal-2024-swift}
Yuze Zhao, Jintao Huang, Jinghan Hu, Xingjun Wang, Yunlin Mao, Daoze Zhang, Zeyinzi Jiang, Zhikai Wu, Baole Ai, Ang Wang, Wenmeng Zhou, and Yingda Chen. 2025.
\newblock \href {https://doi.org/10.1609/aaai.v39i28.35383} {{SWIFT}: A scalable lightweight infrastructure for fine-tuning}.
\newblock In \emph{Proceedings of the Thirty-Ninth AAAI Conference on Artificial Intelligence}, AAAI'25.

\end{thebibliography}

\clearpage

\appendix
\section{Prompt Templates}
\label{app:prompts}

\begin{figure}[h]
\centering
\includegraphics[width=\columnwidth]{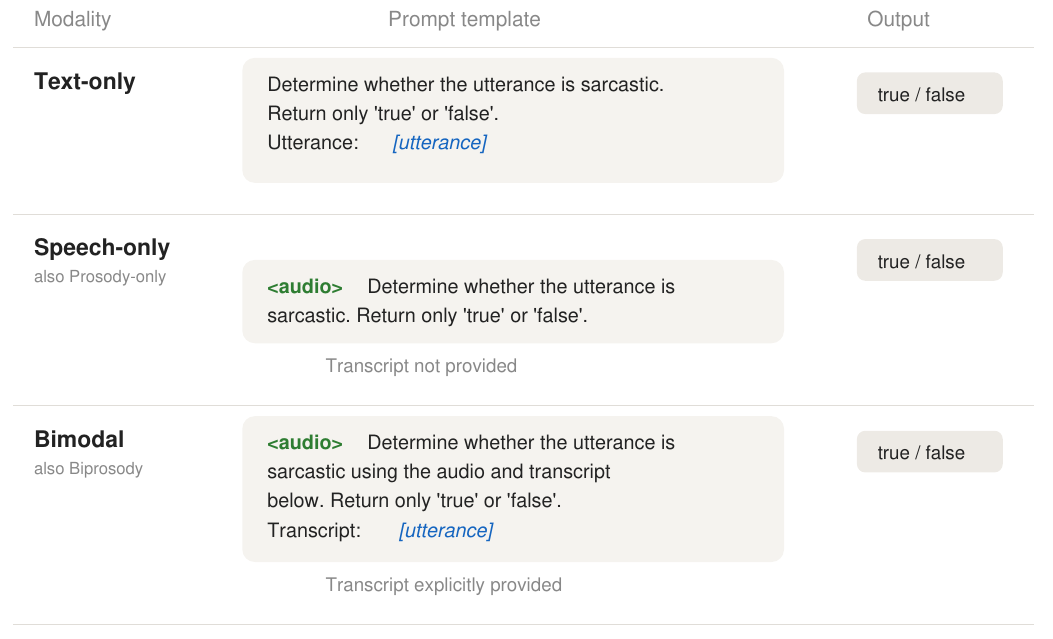}
\caption{Prompt templates for each experimental condition.
\textsc{Speech-only} and \textsc{Prosody-only} share the same
template (audio only); \textsc{Bimodal} and \textsc{Biprosody}
share the same template (audio + transcript). Conditions differ
in whether the audio carries the full voice signal or only
prosodic structure.}
\label{fig:prompts}
\end{figure}

Figure~\ref{fig:prompts} shows the three prompt templates used
across all modality conditions. \textsc{Text-only} receives a
text-only prompt; \textsc{Speech-only} and \textsc{Prosody-only}
share an audio-only prompt; \textsc{Bimodal} and
\textsc{Biprosody} share an audio-with-transcript prompt.
All three templates share the same binary classification
instruction and differ only in whether the input is text,
audio, or audio paired with a transcript, ensuring that any
performance difference between conditions cannot be attributed
to prompt variation.

\section{Speech Enhancement: Full Results}
\label{app:enhancement}
 
\subsection*{Perceptual Quality}
 
\begin{figure}[t]
\centering
\includegraphics[width=0.95\linewidth]{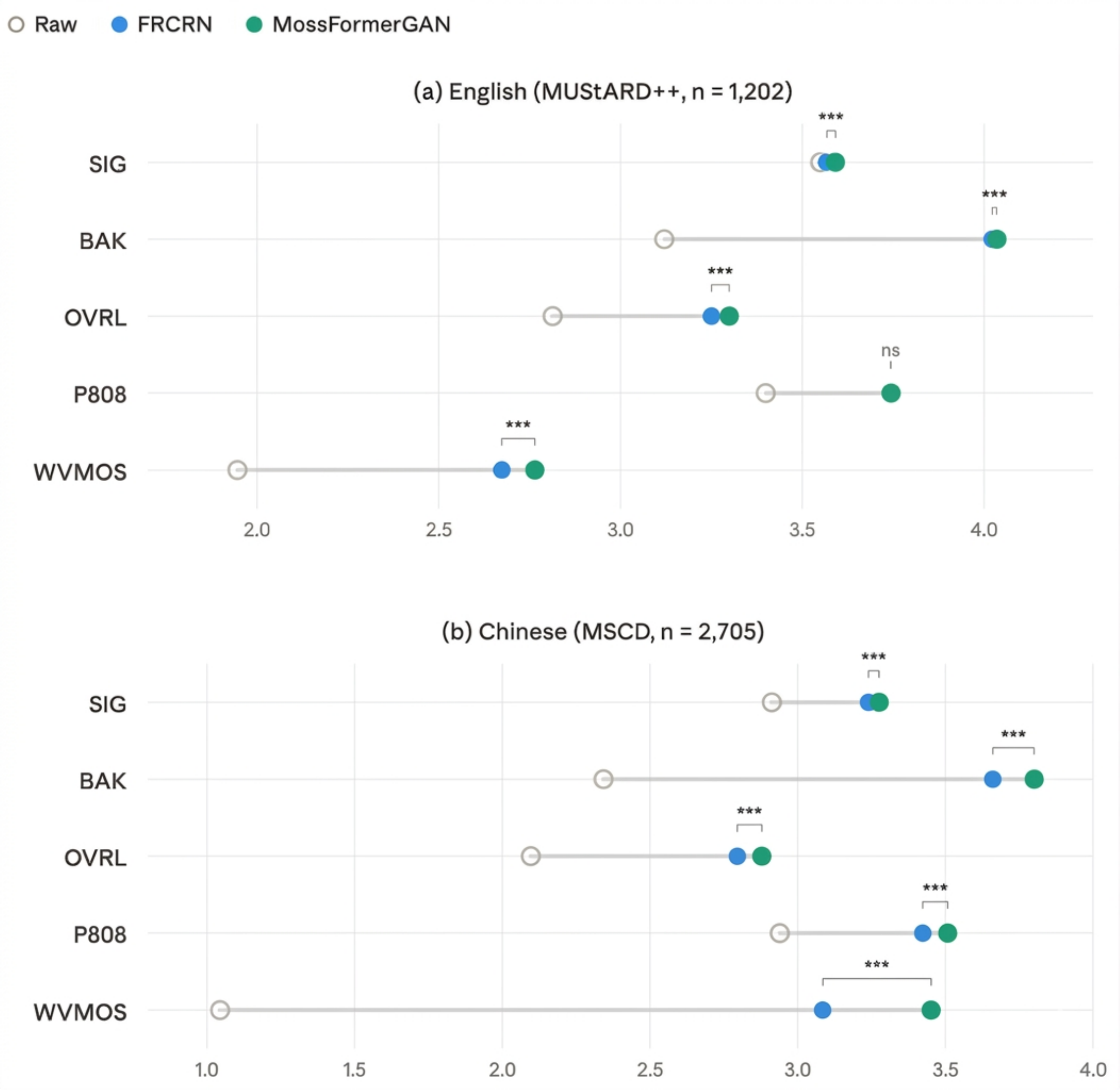}
\caption{Perceptual quality scores (DNSMOS P.835) for English (MUStARD++, $n{=}1{,}202$) and Chinese (MSCD, $n{=}2{,}705$). Hollow circles indicate raw audio; filled circles indicate enhanced audio (blue = FRCRN, green = MossFormerGAN). Connecting lines span from raw to the better-performing model. Both models significantly improve over raw across all metrics (paired $t$-test, $p < 0.001$). Brackets show pairwise significance between FRCRN and MossFormerGAN: {*}{*}{*}\,$p < 0.001$; ns = not significant.}
\label{fig:dnsmos_dumbbell}
\end{figure}

Figure~\ref{fig:dnsmos_dumbbell} presents the full DNSMOS P.835, P808, and WVMOS scores comparing raw audio against FRCRN and MossFormerGAN enhancements. Both models significantly improve over raw audio across all metrics (paired $t$-test, $p < 0.001$). MossFormerGAN further outperforms FRCRN on all metrics ($p < 0.001$), except P808 on English, where the difference is not significant ($p = 0.233$).
 
\subsection*{Raw vs.\ Enhanced: Downstream Speech-Only F1}
 
\begin{figure*}[t]
\centering
\includegraphics[width=0.95\linewidth]{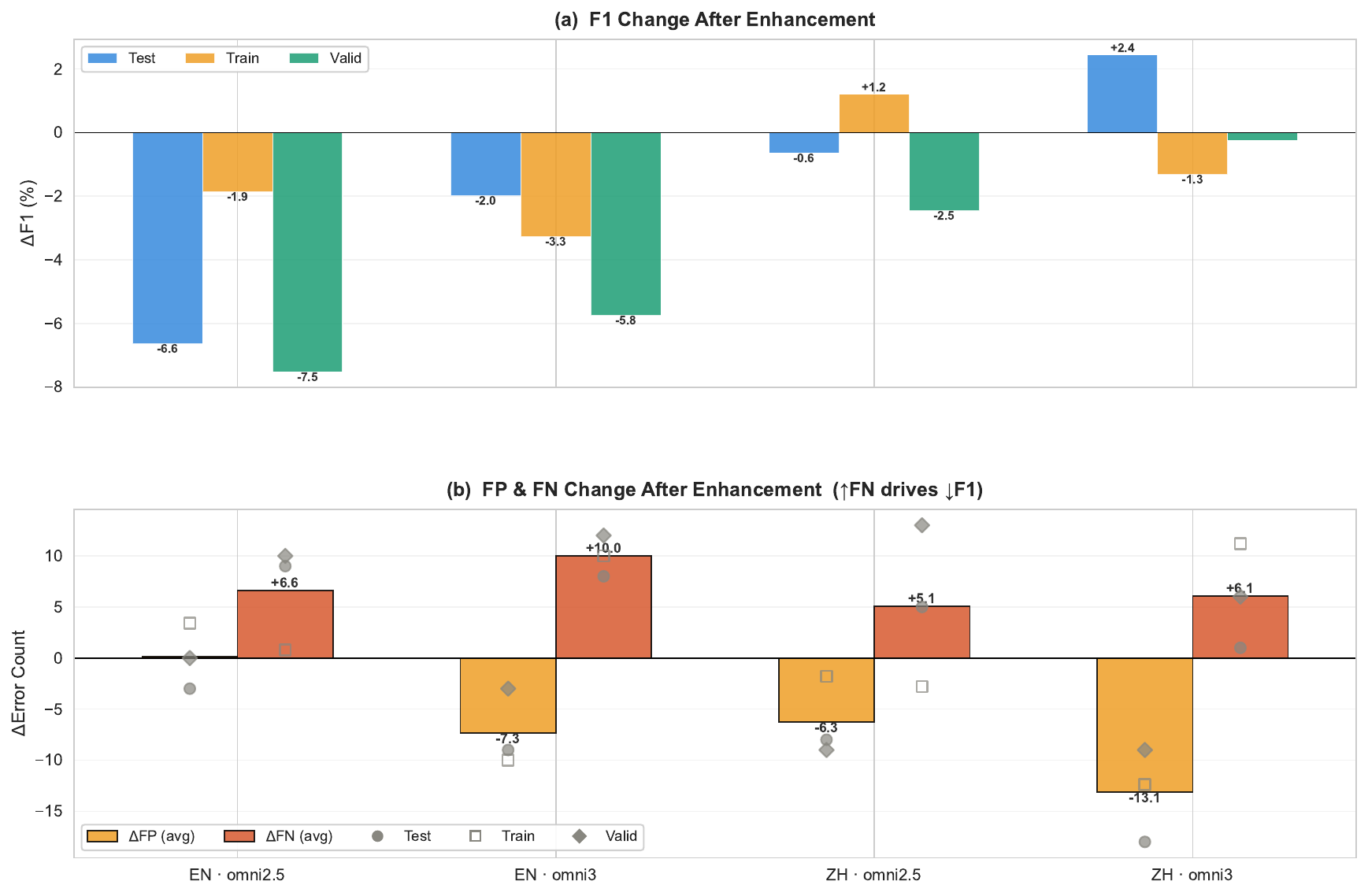}
\caption{Downstream impact of speech enhancement on sarcasm detection. (a)~F1 change after enhancement across splits and conditions. (b)~FP and FN count changes (mean bars with individual split markers); $\uparrow$FN drives $\downarrow$F1.}
\label{fig:raw_vs_enhanced}
\end{figure*}

Although MossFormerGAN improves perceptual quality, we examine whether this translates to downstream gains in the speech-only sarcasm detection setting. Figure~\ref{fig:raw_vs_enhanced} compares speech-only F1 on raw vs.\ MossFormerGAN-enhanced audio. Panel~(a) shows the F1 change (Enhanced $-$ Raw) across train, validation and test splits for each language--model combination. Enhancement generally decreases F1, with the largest drops on English. Panel~(b) decomposes this into false positive and false negative changes: enhancement consistently reduces FPs (orange bars below zero) but increases FNs (red bars above zero), indicating that the enhanced audio trades missed sarcasm detections for fewer false alarms. The net effect is an F1 decrease driven by the FN increase.
 
\section{Per-Split Results}
\label{app:splits}

\begin{figure*}[t]
\centering
\includegraphics[width=\textwidth]{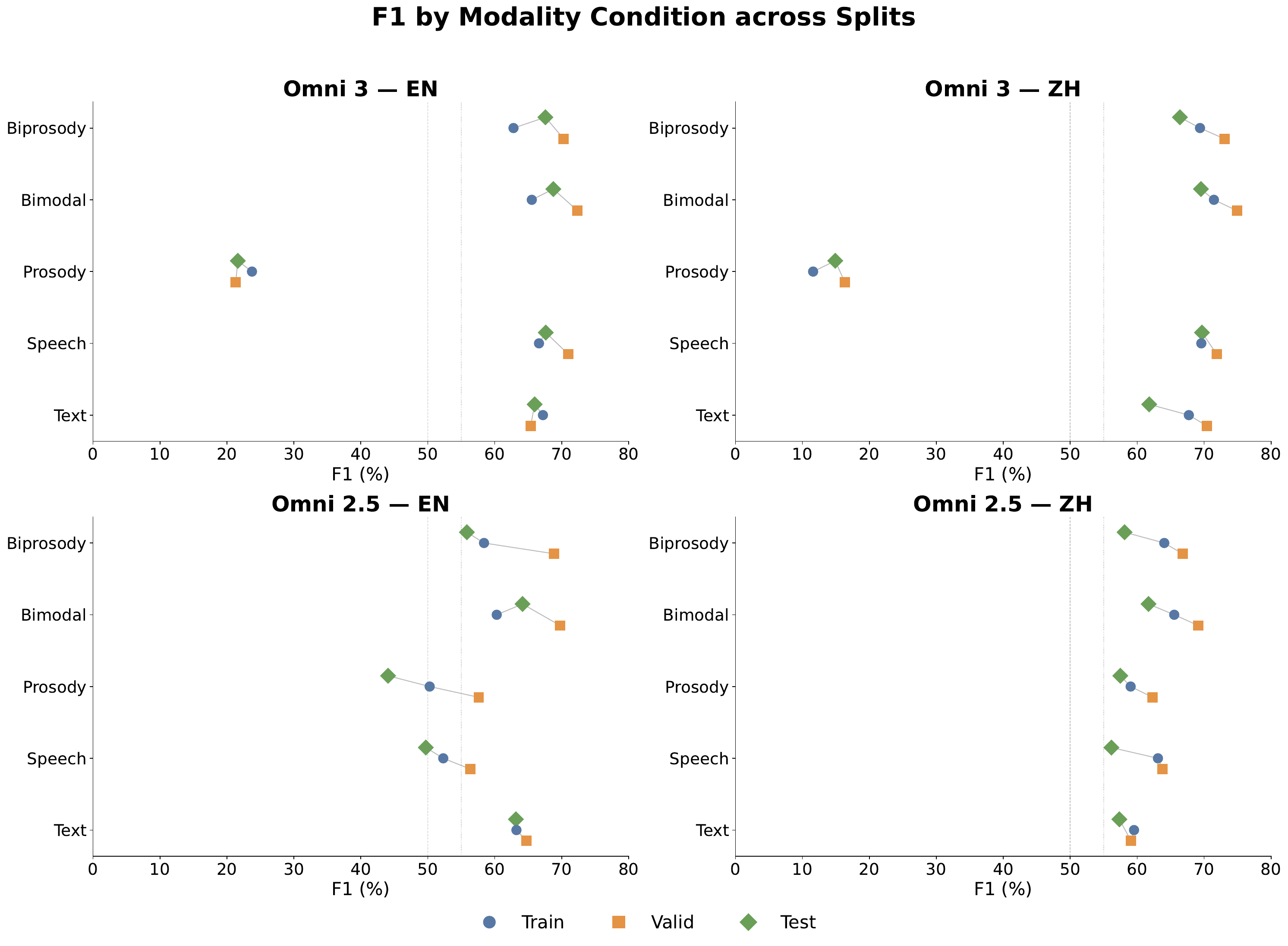}
\caption{F1 (\%) by modality condition for both models and
languages. Each condition shows three points (train, valid,
test); tighter clustering indicates greater cross-split
stability. Dashed line marks the 50\,\% chance baseline.
\textsc{Prosody-only} is excluded from the error analysis
in \S\ref{sec:modality:fp}.}
\label{fig:f1-splits}
\end{figure*}

\begin{figure*}[t]
\centering
\includegraphics[width=\textwidth]{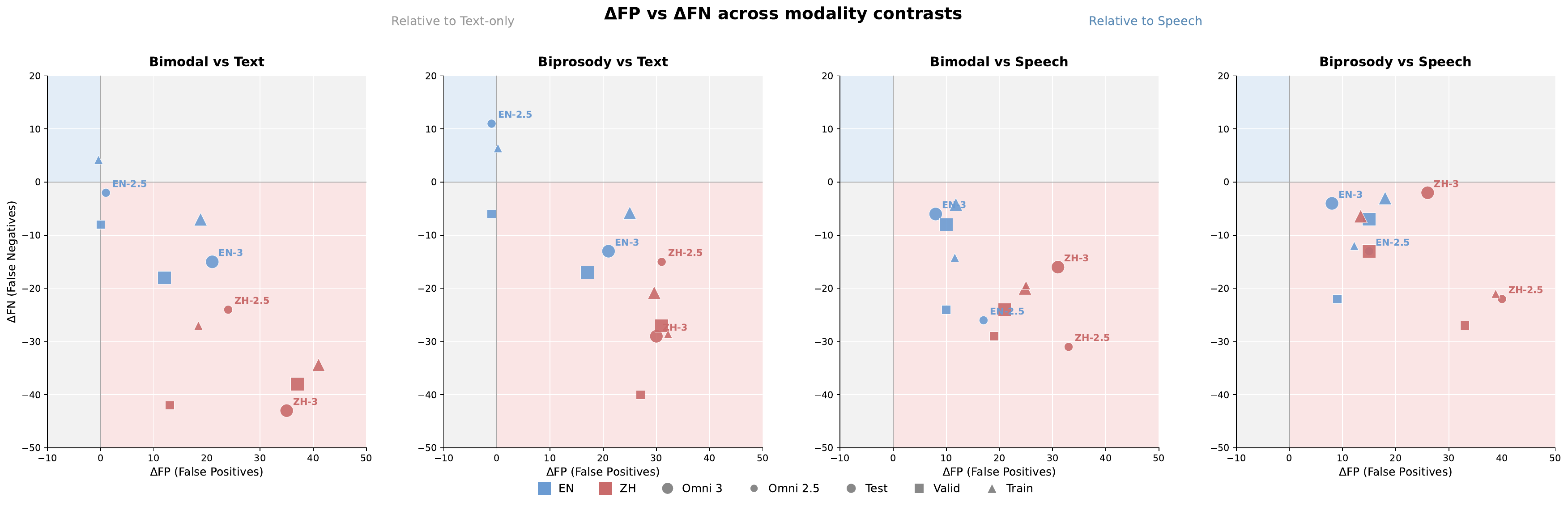}
\caption{$\Delta$FP vs.\ $\Delta$FN for four modality
contrasts across all three splits. Marker shape encodes
split (circle = test, square = valid, triangle = train).
Color encodes language (blue = EN, red = ZH); size encodes
model (larger = Omni~3.0). The lower-right quadrant pattern
replicates consistently across all splits.}
\label{fig:scatter-splits}
\end{figure*}

All models are evaluated zero-shot. The original training split is
partitioned into five folds, which are evaluated independently as repeated evaluation runs rather than for model fitting. Table~\ref{tab:appendix_train_folds} reports the F1 scores for each training fold across all five modality conditions. Figures~\ref{fig:f1-splits} and~\ref{fig:scatter-splits} show
the per-split breakdowns of the main results in
\S\ref{sec:modality:f1} and \S\ref{sec:modality:fp}
respectively. The consistent patterns across the training, validation, and test splits confirm that the observed findings are not artefacts of any particular data partition.

\begin{table*}[t]
\centering
\small
\setlength{\tabcolsep}{4pt}
\begin{tabular}{lllrrrrrr}
\toprule
Language & Model & Modality & Fold 1 & Fold 2 & Fold 3 & Fold 4 & Fold 5 & Mean $\pm$ SD \\
\midrule
\multirow{10}{*}{EN}
& \multirow{5}{*}{Omni 2.5}
  & Text     & 60.96 & 67.72 & 62.56 & 62.83 & 62.11 & 63.24 $\pm$ 2.61 \\
& & Speech   & 47.27 & 56.41 & 54.78 & 50.31 & 52.76 & 52.31 $\pm$ 3.62 \\
& & Prosody  & 54.22 & 45.16 & 52.94 & 47.13 & 51.95 & 50.28 $\pm$ 3.92 \\
& & Bimodal  & 61.20 & 64.48 & 56.83 & 58.64 & 60.32 & 60.29 $\pm$ 2.88 \\
& & Biprosody& 54.05 & 62.11 & 60.34 & 54.95 & 60.54 & 58.40 $\pm$ 3.64 \\
\cmidrule(lr){2-9}
& \multirow{5}{*}{Omni 3}
  & Text     & 66.28 & 68.57 & 66.67 & 67.80 & 66.67 & 67.20 $\pm$ 0.95 \\
& & Speech   & 66.67 & 70.33 & 63.39 & 61.29 & 71.36 & 66.61 $\pm$ 4.33 \\
& & Prosody  & 29.36 & 16.33 & 21.78 & 20.75 & 30.48 & 23.74 $\pm$ 6.01 \\
& & Bimodal  & 63.73 & 68.00 & 63.00 & 63.73 & 69.19 & 65.53 $\pm$ 2.85 \\
& & Biprosody& 59.05 & 66.04 & 64.79 & 63.51 & 60.61 & 62.80 $\pm$ 2.91 \\
\midrule
\multirow{10}{*}{ZH}
& \multirow{5}{*}{Omni 2.5}
  & Text     & 61.11 & 61.39 & 56.63 & 58.25 & 60.29 & 59.53 $\pm$ 2.04 \\
& & Speech   & 60.85 & 64.42 & 62.09 & 66.03 & 62.16 & 63.11 $\pm$ 2.08 \\
& & Prosody  & 59.08 & 55.82 & 60.53 & 58.98 & 60.74 & 59.03 $\pm$ 1.96 \\
& & Bimodal  & 66.67 & 64.77 & 65.16 & 65.81 & 65.30 & 65.54 $\pm$ 0.73 \\
& & Biprosody& 61.17 & 66.81 & 65.65 & 64.26 & 62.37 & 64.05 $\pm$ 2.31 \\
\cmidrule(lr){2-9}
& \multirow{5}{*}{Omni 3}
  & Text     & 67.67 & 71.10 & 65.52 & 67.66 & 66.67 & 67.72 $\pm$ 2.09 \\
& & Speech   & 70.34 & 66.82 & 69.81 & 71.93 & 69.01 & 69.58 $\pm$ 1.88 \\
& & Prosody  &  9.30 & 13.27 &  9.30 & 15.24 & 10.91 & 11.61 $\pm$ 2.60 \\
& & Bimodal  & 73.46 & 70.74 & 70.32 & 73.23 & 69.60 & 71.47 $\pm$ 1.76 \\
& & Biprosody& 67.54 & 70.90 & 67.43 & 72.94 & 68.14 & 69.39 $\pm$ 2.43 \\
\bottomrule
\end{tabular}
\caption{Per-fold F1 scores (\%) on the original training partition
across all five modality conditions. Mean $\pm$ SD is computed across
the five training folds. All models are evaluated zero-shot; the folds
are used as repeated evaluation runs rather than for model fitting.}
\label{tab:appendix_train_folds}
\end{table*}

\section{Acoustic Feature List}
\label{app:features}

Table~\ref{tab:features} lists the 66 acoustic features used in Sections~\ref{sec:acoustic-ground-truth}--\ref{sec:causal}, extracted with Praat/Parselmouth. IOI: inter-onset interval; HNR: harmonics-to-noise ratio; CV: coefficient of variation.

\begin{table}[t]
\centering
\footnotesize
\setlength{\tabcolsep}{4pt}
\begin{tabular}{@{}p{0.28\columnwidth}p{0.66\columnwidth}@{}}
\toprule
\textbf{Category} & \textbf{Features} \\
\midrule
Fundamental frequency (20)
  & mean, median, std, min, max, range, IQR, CV, skewness, kurtosis, voiced fraction; velocity mean/std, acceleration mean/std; num.\ peaks/valleys, peak/valley density, peak-valley ratio \\
\addlinespace[4pt]
Intensity \& energy (17)
  & mean, median, std, min, max, range, CV, skewness, kurtosis; num.\ peaks, peak density, peak prominence mean/std; RMS mean, std, max, range \\
\addlinespace[4pt]
Rhythm \& timing (15)
  & duration, num.\ onsets, speech rate; IOI mean, median, std, CV, range, rhythmic regularity; num.\ pauses, pause rate, pause duration mean/std/total, pause fraction \\
\addlinespace[4pt]
Voice quality (14)
  & HNR mean, median, std, min, max; jitter, shimmer; spectral centroid mean/std, rolloff mean, bandwidth mean, flatness mean; zero-crossing rate mean/std \\
\midrule
\textbf{Total} & \textbf{66 features} \\
\bottomrule
\end{tabular}
\caption{Full acoustic feature set (four categories).}
\label{tab:features}
\end{table}


\section{Manipulation Targets and Actual Outputs}
\label{app:manipulation}

Table~\ref{tab:manipulation} reports the manipulation targets
derived from the FP--Control acoustic profiles (\S\ref{sec:fp-profile})
and the actual PSOLA outputs. $\checkmark$ = on target;
$\approx$ = approximated within acceptable margin;
$\dagger$ = target feature redefined (see main text).

\begin{table*}[tp]
\centering
\small
\begin{tabular}{llccccc}
\toprule
\textbf{Lang.} & \textbf{Feature} & \textbf{FP--Control} & \textbf{Target} & \textbf{Actual} & \textbf{Status} \\
\midrule
\multirow{4}{*}{ZH}
 & \textsc{f0\_mean}         & $+15.6$ Hz ($+8.8\%$)  & $\times1.088$ & $\times1.088$       & \checkmark \\
 & \textsc{f0\_median}       & $+16.0$ Hz ($+9.3\%$)  & $\times1.093$ & $\times1.088$       & $\approx$  \\
 & \textsc{pause\_dur.\ max} & $+0.18$ sec ($+27.1\%$)& $\times1.271$ & $\times1.271$       & \checkmark \\
 & \textsc{pause\_dur.\ std} & $+0.07$ sec ($+41.5\%$)& $\times1.415$ & ${\sim}\times1.415$ & $\approx$  \\
\midrule
\multirow{2}{*}{EN}
 & \textsc{f0\_max}          & $+19.0$ Hz ($+7.1\%$)  & $\times1.071$ & \multirow{2}{*}{$\times1.088$} & \multirow{2}{*}{$\approx$} \\
 & \textsc{f0\_range}        & $+16.1$ Hz ($+10.4\%$) & $\times1.104$ &                     &            \\
 & \textsc{pause duration}   & ---                    & $\times1.490$ & $\times1.490$       & $\dagger$  \\
\bottomrule
\end{tabular}
\caption{Manipulation targets and actual outputs. FP--Control
column shows the observed difference between FP and Control
groups from \S\ref{sec:fp-profile}, which motivates each
manipulation magnitude.
\checkmark\ = on target; $\approx$ = approximated within
acceptable margin; $\dagger$ = target feature redefined
(see text). EN \textsc{f0\_max} and \textsc{f0\_range} share
a single PSOLA factor (averaged $\times1.088$), overshooting
\textsc{f0\_max} and undershooting \textsc{f0\_range} by
comparable margins. EN pause insertion was avoided to preserve
naturalness; existing pauses are instead elongated by
$\times1.49$.}
\label{tab:manipulation}
\end{table*}

\section{Naturalness Validation}
\label{app:naturalness}

Table~\ref{tab:naturalness} reports the full CER and DNSMOS
metrics for manipulated stimuli. CER differences are negligible
for Chinese (0.084 vs.\ 0.079; bootstrap 95\,\% CI crosses zero)
and minimal for English (increase of 0.017). DNSMOS OVRL scores
decrease by 0.30 (EN) and 0.17 (ZH). The weak sample-level CER--MOS correlation
($|r| < 0.15$) confirms that perceptual quality changes do not
systematically impair intelligibility.

\begin{table}[t]
\centering
\small
\setlength{\tabcolsep}{4pt}
\begin{tabular}{lrrrr}
\toprule
& \multicolumn{2}{c}{\textbf{EN (MUStARD++)}}
& \multicolumn{2}{c}{\textbf{ZH (MSCD)}} \\
\cmidrule(lr){2-3}\cmidrule(lr){4-5}
\textbf{Metric} & Clean & Manip. & Clean & Manip. \\
\midrule
CER         & 0.098 & $+0.017^{*}$  & 0.084 & $-0.005^{\phantom{*}}$ \\
DNSMOS OVRL & 2.878 & $-0.302^{*}$  & 3.299 & $-0.173^{*}$ \\
\midrule
\multicolumn{5}{l}{$\Delta$CER--$\Delta$OVRL ($|r|$):
EN $= 0.076$, ZH $= 0.143$} \\
\bottomrule
\end{tabular}
\caption{Clean baseline and change after prosodic manipulation
($n=137$ EN; $n=391$ ZH). Manipulated column shows the
difference from clean (positive CER = degraded intelligibility;
negative OVRL = degraded quality). $^{*}$ denotes significance
(bootstrap 95\,\% CI excludes zero; $p < 0.05$ across sign,
Wilcoxon, and permutation tests). The weak $\Delta$CER--$\Delta$OVRL
correlation confirms quality degradation does not systematically
impair intelligibility at the utterance level.}
\label{tab:naturalness}
\end{table}

\section{Prosodic Feature Ranking (Sarcastic vs.\ Non-Sarcastic)}
\label{app:prosody-ranking}

\begin{figure}[!ht]
\centering
\includegraphics[width=\columnwidth]{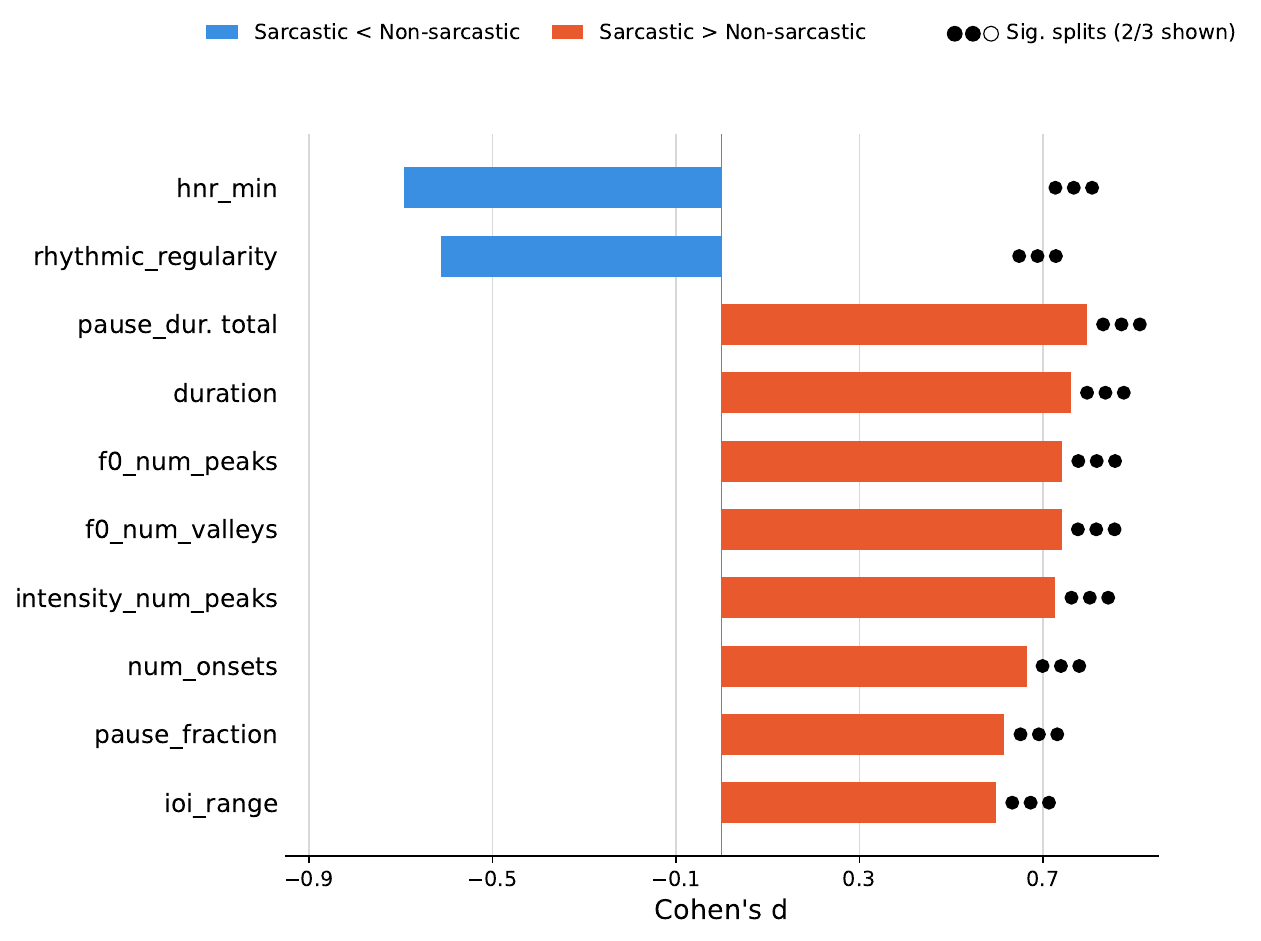}
\caption{Chinese (MSCD): top-10 prosodic features ranked by signed Cohen's $d$. Blue = sarcastic $<$ non-sarcastic; coral = sarcastic $>$ non-sarcastic. All top-10 features significant in 3/3 splits.}
\label{fig:prosody-rank-zh}
\end{figure}

\begin{figure}[!ht]
\centering
\includegraphics[width=\columnwidth]{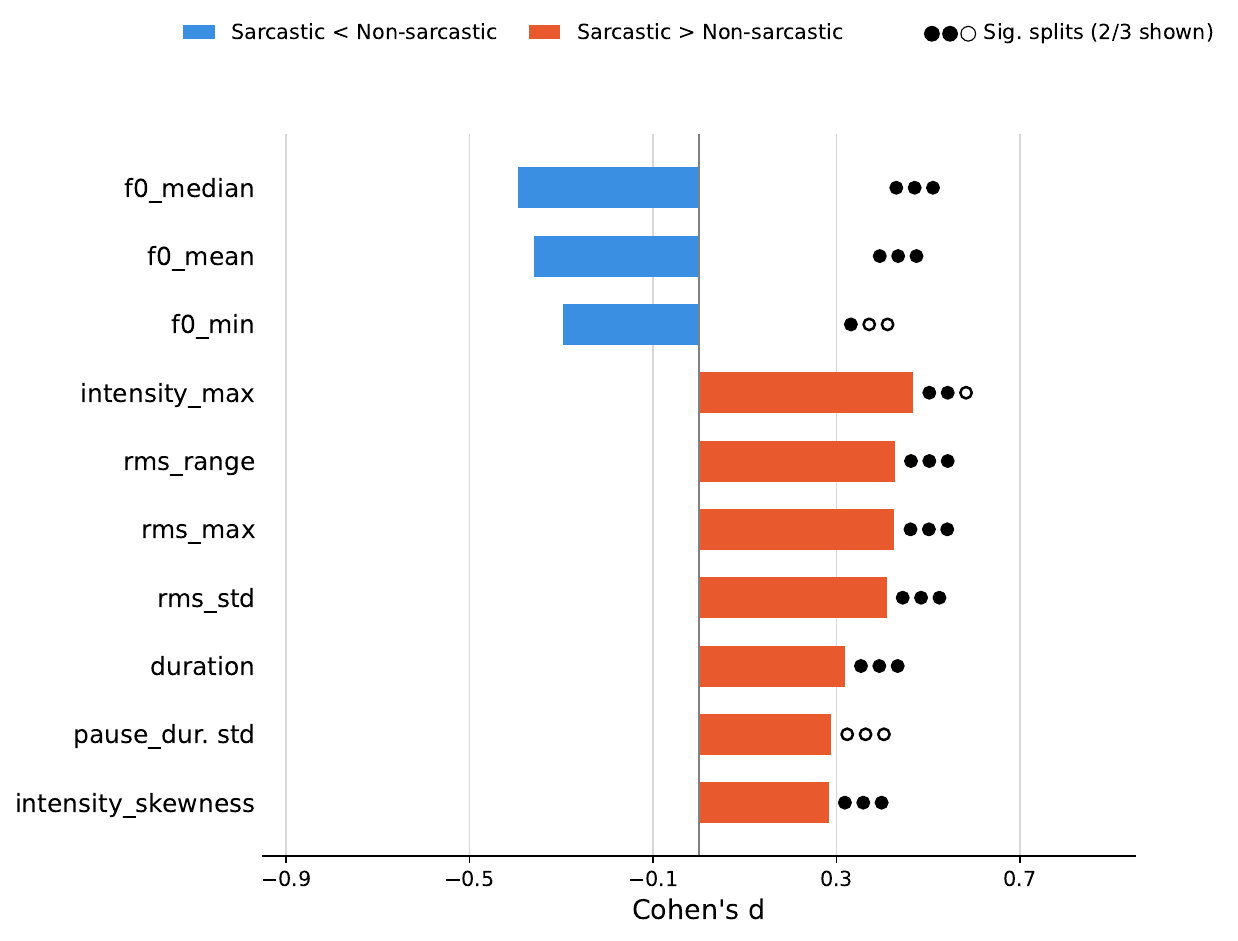}
\caption{English (MUStARD++): top-10 prosodic features ranked by signed Cohen's $d$. Blue = sarcastic $<$ non-sarcastic; coral = sarcastic $>$ non-sarcastic. Dots: filled = significant split ($p < 0.05$), hollow = not (3 dots = 3 splits).}
\label{fig:prosody-rank-en}
\end{figure}

Figures~\ref{fig:prosody-rank-zh} and~\ref{fig:prosody-rank-en} show the top-10 prosodic features distinguishing sarcastic from non-sarcastic utterances, ranked by signed mean Cohen's $d$ across train/valid/test splits (Mann--Whitney $U$). Positive values indicate higher feature values in sarcastic speech; negative values indicate higher values in non-sarcastic speech. Filled dots beside each bar denote the number of splits (out of 3) reaching significance ($p < 0.05$).

For Chinese (Figure~\ref{fig:prosody-rank-zh}), the effect sizes are large and dominated by rhythm and timing features (\texttt{pause\_dur.\ total}, \texttt{duration}, \texttt{num\_onsets}), with all top-10 features reaching significance in 3/3 splits. Sarcastic utterances are longer, contain more pauses, and show reduced rhythmic regularity (\texttt{rhythmic\_regularity}, $d < 0$). The largest negative effect is \texttt{hnr\_min} ($d = -0.69$), indicating that sarcastic speech reaches lower harmonics-to-noise ratios, consistent with moments of breathier or rougher voice quality. 

For English (Figure~\ref{fig:prosody-rank-en}), the strongest discriminators are intensity- and energy-related features (e.g., \texttt{intensity\_max}, \texttt{rms\_range}), with sarcastic utterances exhibiting greater dynamic range. Fundamental frequency features (\texttt{f0\_median}, \texttt{f0\_mean}) show the opposite pattern: sarcastic speech has a lower pitch register overall. Notably, \texttt{pause\_dur.\ std} ranks 7th by absolute effect size but reaches significance in 0/3 splits, suggesting that its effect is unstable across data partitions.

\section{False Positive vs.\ Control Feature Profiles}
\label{app:fp-profile}

\begin{figure}[!ht]
\centering
\includegraphics[width=\columnwidth]{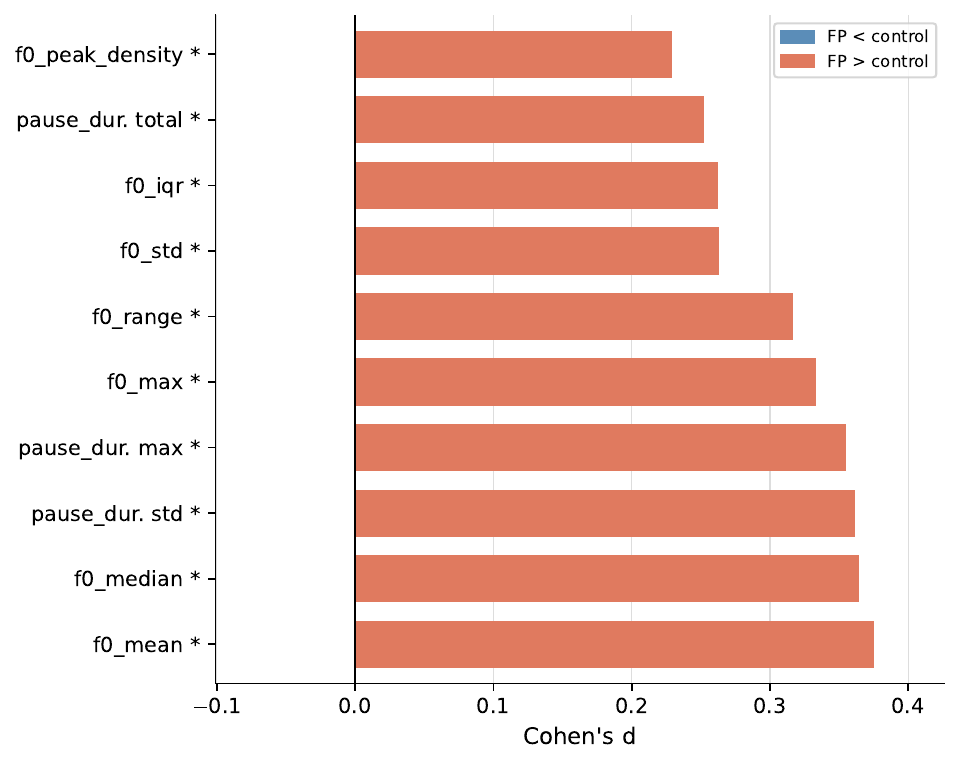}
\caption{Chinese (MSCD): top-10 features separating false positives from matched controls, ranked by signed Cohen's $d$. All features significant ($* = p < 0.05$). FP utterances show elevated pitch and irregular pausing relative to controls.}
\label{fig:fp-profile-zh}
\end{figure}

\begin{figure}[!ht]
\centering
\includegraphics[width=\columnwidth]{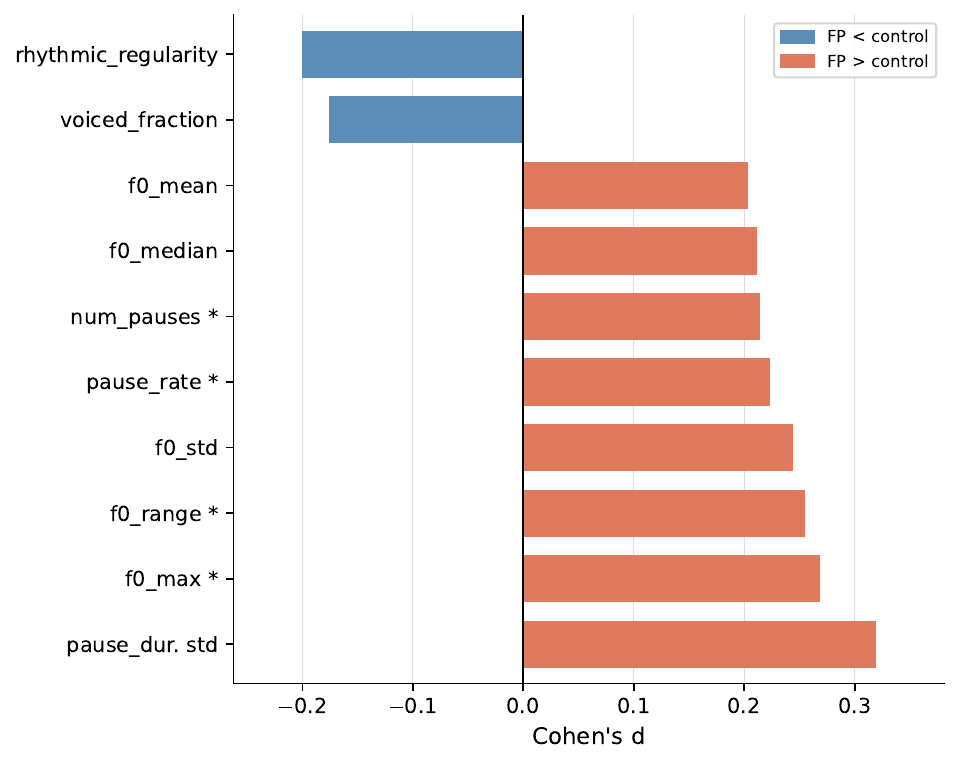}
\caption{English (MUStARD++): top-10 features separating false positives from matched controls, ranked by signed Cohen's $d$. Only 4 features reach significance ($* = p < 0.05$). Coral = FP $>$ Control; blue = FP $<$ Control.}
\label{fig:fp-profile-en}
\end{figure}

Figures~\ref{fig:fp-profile-zh} and~\ref{fig:fp-profile-en} show the top-10 features separating false positive (FP) utterances from matched non-sarcastic controls, ranked by signed Cohen's $d$ (FP $-$ Control). Features marked with * reach significance ($p < 0.05$, Mann--Whitney $U$). Dashed lines indicate $|d| = 0.2$ (small effect threshold).

For Chinese (Figure~\ref{fig:fp-profile-zh}), all 10 features are significant and positively directed, indicating that FP utterances exhibit higher values than controls. The profile is dominated by F0 features (\texttt{f0\_mean}, \texttt{f0\_median}, \texttt{f0\_max}, \texttt{f0\_range}, \texttt{f0\_std}, \texttt{f0\_iqr}, \texttt{f0\_peak\_density}) and pause irregularity (\texttt{pause\_duration\_std}, \texttt{pause\_duration\_max}). This contrasts with the ground-truth sarcasm signature (Appendix~\ref{app:prosody-ranking}), where temporal features such as \texttt{pause\_duration\_total} and \texttt{duration} dominate.

For English (Figure~\ref{fig:fp-profile-en}), only 4 of 10 features reach significance (\texttt{f0\_max}, \texttt{f0\_range}, \texttt{pause\_rate}, \texttt{num\_pauses}). The significant features are all positively directed, with elevated pitch extremes and increased pausing. The increased F0 statistics are directionally opposite to the ground-truth English sarcasm profile, where F0 features are negative (sarcastic speech has lower pitch). Two negative features (\texttt{rhythmic\_regularity}, \texttt{voiced\_fraction}) suggest FP utterances are also less rhythmically regular and less continuously voiced than controls.

\section{Reverse Manipulation Results}
\label{app:reverse}

To provide bidirectional causal evidence, we shifted FP samples
toward control-like prosodic targets by reversing the
manipulation directions applied in \S\ref{sec:causal:design}.
Table~\ref{tab:reverse} reports the correction rates, defined
as the fraction of FP samples whose prediction flips to
non-sarcastic following the reverse manipulation. Partial
recovery is expected given that FP samples may carry additional
acoustic or contextual features beyond the two targeted
dimensions.

\begin{table}[h]
\centering
\small
\setlength{\tabcolsep}{4pt}
\begin{tabular}{lcccc}
\toprule
 & \multicolumn{2}{c}{\textbf{EN}} & \multicolumn{2}{c}{\textbf{ZH}} \\
\cmidrule(lr){2-3}\cmidrule(lr){4-5}
\textbf{Modality} & Omni 2.5 & Omni 3 & Omni 2.5 & Omni 3 \\
\midrule
\textsc{Speech-only} & 56.3 & 56.3 & 53.9 & 53.9 \\
\textsc{Bimodal}     & 45.1 & 31.0 & 38.9 & 35.2 \\
\textsc{Biprosody}   & 53.5 & 33.8 & 29.5 & 38.9 \\
\bottomrule
\end{tabular}
\caption{Reverse manipulation correction rates (\%): fraction
of FP samples that flip to correct non-sarcastic classification
after shifting prosodic features toward control-like targets.}
\label{tab:reverse}
\end{table}

\section{Gemini~3 Flash Preview: Extended Results}
\label{app:gemini}

The Omni-derived manipulation template, which targets elevated
F0 and irregular pausing, transfers to Gemini~3 Flash Preview
with non-trivial flip rates across all conditions
(\S\ref{sec:causal:transfer}), establishing that the prosodic
heuristic is not confined to the Qwen MLLMs. To
characterise Gemini~3's own FP acoustic profile, we applied
the same FP--Control acoustic analysis described in
\S\ref{sec:fp-profile} and derived a language-specific
manipulation template from the resulting geometry.

Gemini~3's FP--Control geometry reveals that the prosodic
heuristic operates through partially distinct acoustic
dimensions across model families. For Chinese, FP utterances
are characterised by higher RMS energy, intensity variability,
and F0 contour activity, dimensions not targeted by the
Omni-derived template, which explains the lower Chinese
transfer rates. For English, the FP profile involves elevated
central pitch, reduced F0 contour complexity, and slower
temporal structure, partially overlapping with the Omni
template and consistent with the higher English transfer
rates observed in \S\ref{sec:causal:transfer}. The
Gemini-specific template, which directly targets these
dimensions, produces higher flip rates across both conditions
and languages, with the largest gain in English
\textsc{Speech-only} ($\Delta = +21.0$\,pp), confirming that
the heuristic is operative in Gemini~3 and that the
cross-model transfer rates reflect a partial rather than
complete instantiation of the same vulnerability.

\begin{table}[t]
\centering
\small
\setlength{\tabcolsep}{5pt}
\begin{tabular}{llrr}
\toprule
& & \multicolumn{2}{c}{\textbf{Flip Rate (\%)}} \\
\cmidrule(lr){3-4}
\textbf{Condition} & \textbf{Lang.} & \textbf{Omni} &
\textbf{Gemini ($\Delta$)} \\
\midrule
\multirow{2}{*}{\textsc{Bimodal}}
  & EN & 14.3 & $+5.0$ \\
  & ZH &  5.0 & $+1.2$ \\
\midrule
\multirow{2}{*}{\textsc{Speech-only}}
  & EN &  8.9 & $+21.0$ \\
  & ZH & 10.5 & $+0.0$  \\
\bottomrule
\end{tabular}
\caption{FP flip rates (\%) for Gemini~3 Flash Preview under
the Omni-derived template (Omni) and the change when using
the Gemini-specific template ($\Delta$ = Gemini $-$ Omni).
Positive values indicate higher flip rates under the
Gemini-specific template.}
\label{tab:gemini-templates}
\end{table}

\section{Chain-of-Thought Reasoning Traces}
\label{app:cot-examples}

Omni~3.0 operates in thinking mode, producing explicit chain-of-thought reasoning prior to the final binary prediction. We present two representative examples illustrating distinct failure mechanisms.

\subsection*{Example 1: Mismatch Heuristic (English)}

\begin{figure*}[t]
\centering
\includegraphics[width=0.95\linewidth]{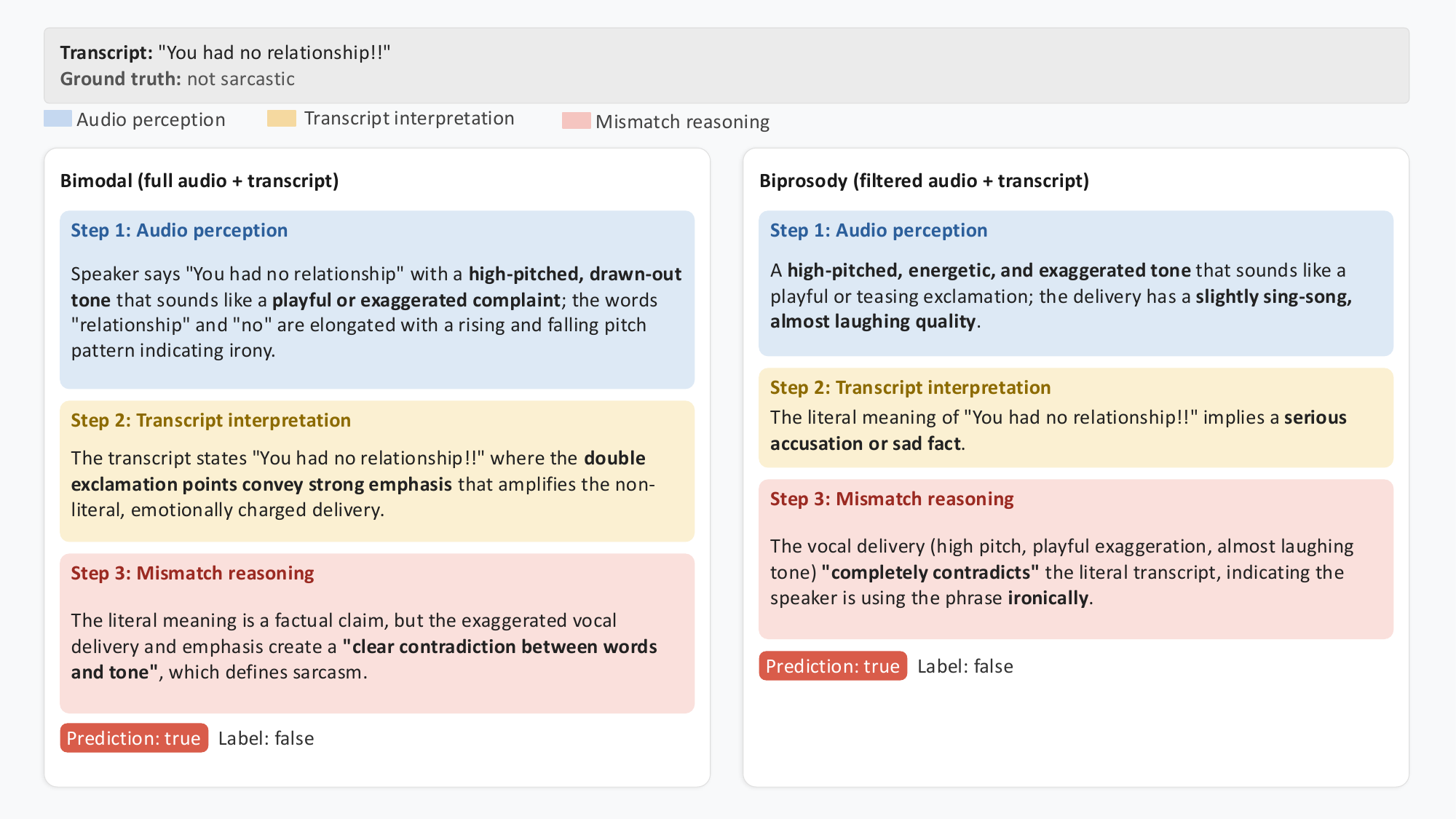}
\caption{Example~1: Chain-of-thought traces for a representative false positive under \textsc{Bimodal} and \textsc{Biprosody}. Both traces follow an identical three-step pattern (audio perception, transcript interpretation, mismatch reasoning) and arrive at the same incorrect sarcasm prediction. Key phrases bolded.}
\label{fig:cot-mismatch}
\end{figure*}

Figure~\ref{fig:cot-mismatch} presents the reasoning traces for a false positive (ground-truth label: \textit{not sarcastic}, transcript: \textit{``You had no relationship!!''}) under both the \textsc{Bimodal} and \textsc{Biprosody} conditions. Both traces follow the same three-step pattern: (1)~the model characterises the audio as having exaggerated, playful prosody; (2)~it interprets the transcript as conveying serious or factual content; (3)~it concludes that the perceived contrast between tone and text constitutes sarcastic irony. The \textsc{Biprosody} trace is particularly revealing: despite receiving audio from which lexical content has been removed, the model still constructs a detailed narrative about ``playful exaggeration'' and ``almost laughing quality,'' attributing these characteristics to prosodic contour alone. The convergence of both traces on the same mismatch conclusion confirms that the false positive is driven by perceived text--prosody incongruence rather than any information recoverable from vocal semantics.

\subsection*{Example 2: Modality Context Effect}

\begin{figure*}[t]
\centering
\includegraphics[width=0.95\linewidth]{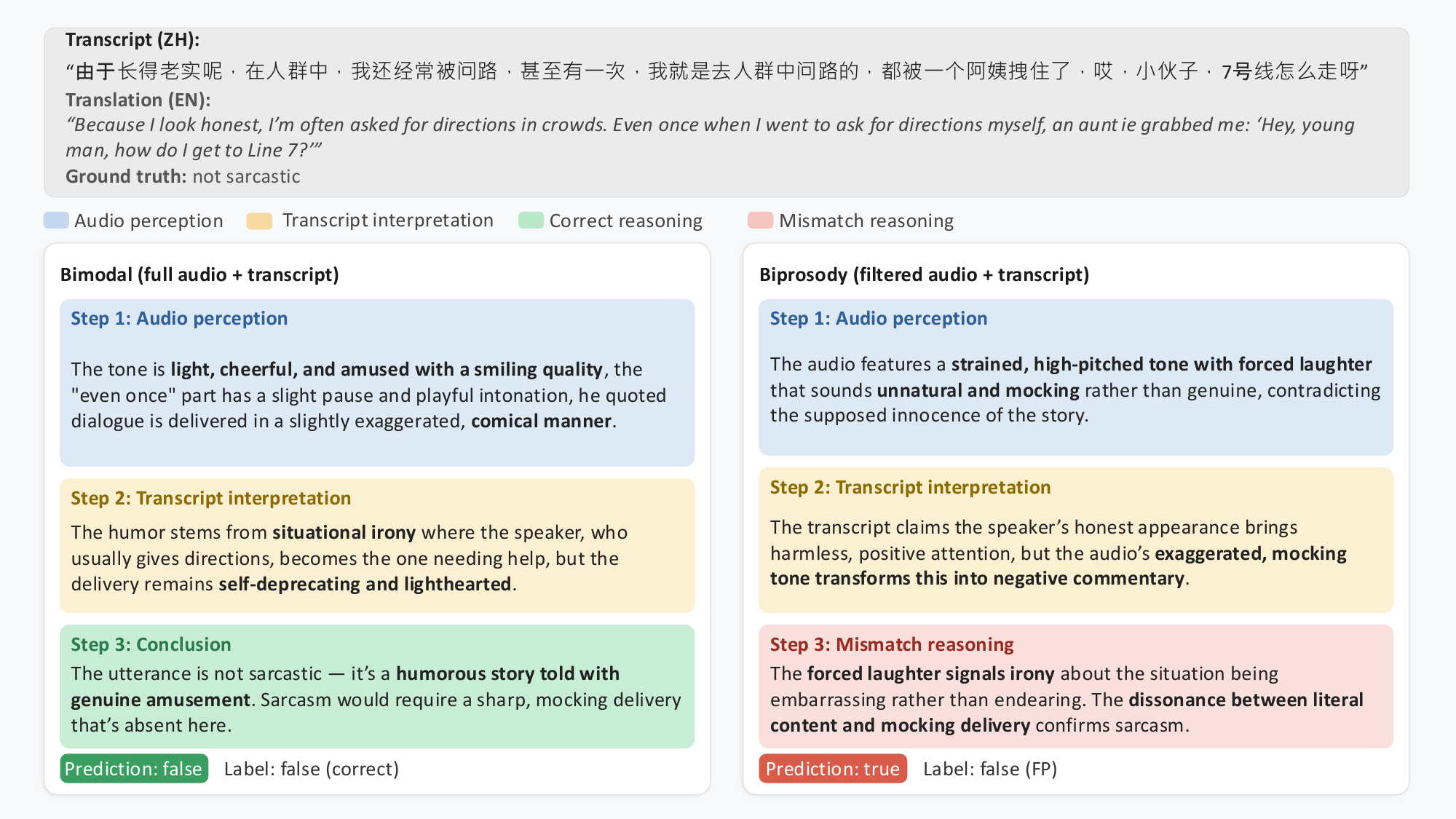}
\caption{Example~2: Chain-of-thought traces for a causally manipulated Chinese sample ($+8.8\%$ F0) under \textsc{Bimodal} (correct rejection) and \textsc{Biprosody} (false positive). Key phrases bolded.}
\label{fig:cot-causal}
\end{figure*}

Figure~\ref{fig:cot-causal} presents the traces for a causally manipulated Chinese sample. Under \textsc{Bimodal}, the model correctly identifies the utterance as a humorous anecdote told with genuine amusement, describing the tone as ``light, cheerful, amused with a smiling quality'' and predicting \textit{false}. Under \textsc{Biprosody}, the identical prosodic manipulation is instead described as ``strained, high-pitched'' with ``forced laughter that sounds unnatural and mocking,'' leading the model to conclude that the delivery contradicts the transcript and predict \textit{true} (false positive).

This divergence illustrates the modality context effect: in \textsc{Bimodal}, the acoustic content of full speech provides a grounding context that normalises the manipulated prosodic features, while in \textsc{Biprosody}, the absence of vocal semantics causes the model to interpret identical prosody as mocking rather than expressive. Prosodic features do not carry fixed perceptual weight for these models but are interpreted relative to the broader acoustic context in which they occur~\citep{corrêa2025emotion, sun2025one}.

\end{document}